\documentclass[journal]{IEEEtran}
\PassOptionsToPackage{dvipsnames}{xcolor}

\ifCLASSOPTIONcompsoc
  \usepackage[nocompress]{cite}
\else
  \usepackage{cite}
\fi

\ifCLASSINFOpdf
  \usepackage[pdftex]{graphicx}
  \DeclareGraphicsExtensions{.pdf,.jpeg,.png}
\else
  \usepackage[dvips]{graphicx}
  \DeclareGraphicsExtensions{.eps}
\fi

\ifCLASSOPTIONcompsoc
  \usepackage[caption=false,font=footnotesize,labelfont=sf,textfont=sf]{subfig}
\else
  \usepackage[caption=false,font=footnotesize]{subfig}
\fi

\usepackage{ragged2e}
\usepackage{lipsum}
\usepackage{comment}
\usepackage{color,soul}
\usepackage{booktabs}
\usepackage{multirow}
\usepackage{tabularx}
\newcolumntype{C}{>{\centering\arraybackslash}X}
\usepackage{amsmath}
\usepackage{bm}
\usepackage{url}            % simple URL typesetting
\usepackage{mathtools}
\DeclarePairedDelimiter{\norm}{\lVert}{\rVert}

\usepackage{tikz,tikz-3dplot}
\usetikzlibrary{calc}
\usetikzlibrary{shapes,positioning,intersections,quotes}
\usepackage{graphicx}
\usepackage{footmisc}
\usepackage{booktabs}
\usepackage{amsmath}
\usepackage{amsfonts}
\usepackage{amsbsy}
\usepackage{placeins}
\usepackage{siunitx}
\usepackage{bm}
\usepackage[utf8]{inputenc} % allow utf-8 input
\usepackage[T1]{fontenc}    % use 8-bit T1 fonts
\usepackage{url}            % simple URL typesetting
\usepackage{nicefrac}       % compact symbols for 1/2, etc.
\usepackage{multirow}
\usepackage{mathtools}
\usepackage{xparse}
\usepackage{empheq}

\usepackage[switch]{lineno}
\usepackage{centernot}
\usepackage{xcolor}

\newcommand\blfootnote[1]{%
  \begingroup
  \renewcommand\thefootnote{}\footnote{#1}%
  \addtocounter{footnote}{-1}%
  \endgroup
}
\begin{document}
%
%\title{Entropic Out-of-Distribution Detection}
\title{Entropic Out-of-Distribution Detection:\\Seamless Detection of Unknown Examples}

\begin{comment}
\author{
David~Mac\^edo,~\IEEEmembership{Member,~IEEE,}
Tsang~Ing~Ren,~\IEEEmembership{Member,~IEEE,}
Cleber~Zanchettin,~\IEEEmembership{Member,~IEEE,}
Adriano~L.~I.~Oliveira,~\IEEEmembership{Senior Member,~IEEE,}
and~Teresa~Ludermir,~\IEEEmembership{Senior Member,~IEEE,}% <-this % stops a space
\IEEEcompsocitemizethanks{
\IEEEcompsocthanksitem David~Mac\^edo, Tsang~Ing~Ren, Cleber~Zanchettin, Adriano~L.~I.~Oliveira, and~Teresa~Ludermir are with the Centro de Inform\'atica, Universidade Federal de Pernambuco, Brazil. E-mail: dlm@cin.ufpe.br. David~Mac\^edo was with Montreal Institute for Learning Algorithms (MILA), 
Université de Montréal (UdeM), Quebec, Canada.
% David~Mac\^edo was with Montreal Institute for Learning Algorithms, University of Montreal, Quebec, Canada.
%David~Mac\^edo was with Montreal Institute for Learning Algorithms, University of Montreal, Quebec, Canada.
\protect\\
% note need leading \protect in front of \\ to get a newline within \thanks as
% \\ is fragile and will error, could use \hfil\break instead.
%E-mail: see dlm@cin.ufpe.br
%\IEEEcompsocthanksitem David~Mac\^edo was with Montreal Institute for Learning Algorithms, University of Montreal, Quebec, Canada.
}% <-this % stops an unwanted space
%\thanks{Manuscript received August 1, 2020; revised November 1, 2020.}
}
\end{comment}

\author{
David~Mac\^edo,~\IEEEmembership{Member,~IEEE,}
Tsang~Ing~Ren,~\IEEEmembership{Member,~IEEE,}
Cleber~Zanchettin,~\IEEEmembership{Member,~IEEE,}
Adriano~L.~I.~Oliveira,~\IEEEmembership{Senior Member,~IEEE,}
and~Teresa~Ludermir,~\IEEEmembership{Senior Member,~IEEE}% <-this % stops a space
\thanks{David~Mac\^edo, Tsang~Ing~Ren, Cleber~Zanchettin, Adriano~L.~I.~Oliveira, and~Teresa~Ludermir are with the Centro de Inform\'atica, Universidade Federal de Pernambuco, Brazil. E-mail: dlm@cin.ufpe.br.}% <-this % stops a space
\thanks{David~Mac\^edo was with Montreal Institute for Learning Algorithms (MILA), 
Université de Montréal (UdeM), Quebec, Canada.}% <-this % stops a space
}

%\begin{comment}
\markboth{IEEE TRANSACTIONS ON NEURAL NETWORKS AND LEARNING SYSTEMS: SPECIAL ISSUE ON DEEP LEARNING FOR ANOMALY DETECTION}{Mac\^edo~\MakeLowercase{\textit{et al.}}: Entropic Out-of-Distribution Detection: Seamless Detection of Unknown Examples}
%\end{comment}

% make the title area
\maketitle % COMMENT FOR TPAMI

%\IEEEtitleabstractindextext{% UNCOMMENT FOR TPAMI
\begin{abstract}
In this paper, we argue that the unsatisfactory out-of-distribution (OOD) detection performance of neural networks is mainly due to the SoftMax loss anisotropy and propensity to produce low entropy probability distributions in disagreement with the principle of maximum entropy. Current out-of-distribution (OOD) detection approaches usually do not directly fix the SoftMax loss drawbacks, but rather build techniques to circumvent it. Unfortunately, those methods usually produce undesired side effects (e.g., classification accuracy drop, additional hyperparameters, slower inferences, and collecting extra data). In the opposite direction, we propose replacing SoftMax loss with a novel loss function that does not suffer from the mentioned weaknesses. The proposed IsoMax loss is isotropic (exclusively distance-based) and provides high entropy posterior probability distributions. Replacing the SoftMax loss by IsoMax loss requires no model or training changes. Additionally, the models trained with IsoMax loss produce as fast and energy-efficient inferences as those trained using SoftMax loss. Moreover, no classification accuracy drop is observed. The proposed method does not rely on outlier/background data, hyperparameter tuning, temperature calibration, feature extraction, metric learning, adversarial training, ensemble procedures, or generative models. Our experiments showed that IsoMax loss works as a seamless SoftMax loss drop-in replacement that significantly improves neural networks' OOD detection performance. Hence, it may be used as a baseline OOD detection approach to be combined with current or future OOD detection techniques to achieve even higher results.
\end{abstract}

% Note that keywords are not normally used for peerreview papers.
%}% UNCOMMENT FOR TPAMI

% make the title area
%\maketitle % UNCOMMENT FOR TPAMI

%\IEEEdisplaynontitleabstractindextext % UNCOMMENT FOR TPAMI
\IEEEpeerreviewmaketitle
%\IEEEraisesectionheading{
\section{Introduction}\label{sec:introduction}
%}% UNCOMMENT FOR TPAMI

%\begin{comment}
\begin{table*}%[!ht]
\setlength{\tabcolsep}{2pt}
%\small
\renewcommand{\arraystretch}{0.6}
%\caption{\normalsize \textcolor{red}{Out-of-Distribution Detection: Approaches, Techniques, and Side Effects.}}
\caption{Out-of-Distribution Detection: Approaches, Techniques, and Side Effects.}
\vskip -0.2cm
%\vskip -0.2cm
%\vspace{-0.1in}
\label{tab:overview}
\centering
%\begin{tabularx}{\textwidth}{clCCC}
\begin{tabularx}{\textwidth}{l|CCCCC}
\toprule
%& & & & &\\
Techniques and Side Effects & \mbox{SoftMax Loss} (current baseline) & ODIN \cite{liang2018enhancing} & Mahalanobis \cite{lee2018simple} & ACET \cite{Hein2018WhyRN} & IsoMax Loss (proposed baseline)\\
%& & & & &\\
\midrule
Inference input preprocessing: & & & &\\
Multiple forward passes and backpropagation. & \color{blue}Not Required & \color{red}Required & \color{red}Required & \color{blue}Not Required & \color{blue}Not Required\\
At least three times slower inference. & \color{blue}(Fast) & \color{red}(Slow) & \color{red}(Slow) & \color{blue}(Fast) & \color{blue}(Fast)\\
\bf{Slow inferences.} & & & &\\
\midrule%{2-5}
Inference input preprocessing: & & & &\\
At least three times higher energy consumption. & \color{blue}Not Required & \color{red}Required & \color{red}Required & \color{blue}Not Required & \color{blue}Not Required\\
At least three times higher computational cost. & \color{blue}(Efficient) & \color{red}(Not Efficient) & \color{red}(Not Efficient) & \color{blue}(Efficient) & \color{blue}(Efficient)\\
\bf{Energy-inefficient inferences.} & & & &\\
\midrule%{2-5}
%Feature Ensemble: & & & &\\
%Low-Level Features Dependency. & \color{blue}Not Required & \color{blue}Not Required & \color{red}Required & \color{red}Required & \color{blue}Not Required\\
%Adversarial Training or Adversarial Examples Generation.  & \color{blue}(Scalable) & \color{blue}(Scalable) & \color{red}(Not Scalable) & \color{red}(Not Scalable) & \color{blue}(Scalable)\\
%LIMITED SCALABILITY & & & &\\
Adversarial training: & & & &\\
Previously known adversarial hyperparameters. & \color{blue}Not Required & \color{blue}Not Required & \color{blue}Not Required & \color{red}Required & \color{blue}Not Required\\
Slower and more hyperparametrized training.  & \color{blue}(Scalable) & \color{blue}(Scalable) & \color{blue}(Scalable) & \color{red}(Not Scalable) & \color{blue}(Scalable)\\
\bf{Limited scalability.} & & & &\\
\midrule%{2-5}
%Hyperparameter Validation or Metric Learning:  & & & &\\
%Out-of-Distribution or Adversarial Examples. & \color{blue}Not Required & \color{red}Required & \color{red}Required & \color{red}Required & \color{blue}Not Required\\
%Adversarial Training Hyperparameters. & \color{blue}(Turnkey) & \color{red}(Not Turnkey)& \color{red}(Not Turnkey) & \color{red}(Not Turnkey) & \color{blue}(Turnkey)\\
%ADDITIONAL PROCEDURES & & & &\\
Adversarial validation:  & & & &\\
Previously known adversarial hyperparameters. & \color{blue}Not Required & \color{red}Required & \color{red}Required & \color{blue}Not Required & \color{blue}Not Required\\
Adversarial examples generation. & \color{blue}(Turnkey) & \color{red}(Not Turnkey)& \color{red}(Not Turnkey) & \color{blue}(Turnkey) & \color{blue}(Turnkey)\\
\bf{Hyperparameters tuning.} & & & &\\
\bottomrule
\end{tabularx}
%\vskip -0.5cm
\begin{justify}IsoMax loss has no drawbacks compared with SoftMax loss while presenting higher OOD detection performance (Table~\ref{tbl:expanded_fair_odd}). ODIN, the Mahalanobis method, and ACET present weaknesses from a practical use perspective. However, if these limitations are not a concern for an application, they may be combined with IsoMax loss to improve overall OOD detection performance. The strengths of the competing approaches are in blue, while weaknesses are in red.\end{justify}
\end{table*}
%\end{comment}

\IEEEPARstart{N}{eural} networks have been used as classifiers in a wide range of applications. Their design usually considers that the model receives an instance of one of the trained classes at inference. If this holds, neural networks commonly present satisfactory performance. However, in real-world applications, this assumption may not be fulfilled.% In such cases, neural networks present overconfident predictions even for objects they were not trained to recognize \cite{Guo2017OnCO}.

The ability to detect whether an input applied to a neural network does not represent an example of the trained classes is essential to applications in medicine, finance, agriculture, business, marketing, and engineering. In such situations, it is better to have a system able to acknowledge that the sample should not be classified. The rapid adoption of neural networks in modern applications makes the development of such capability a primary necessity from a practical point of view. This problem has been studied under many similar nomenclatures, such as open set recognition \cite{Scheirer_2013_TPAMI,Scheirer_2014_TPAMIb} and open-world recognition \cite{7298799,Rudd_2018_TPAMI}.

Recently, Hendrycks and Gimpel \cite{hendrycks2017baseline} defined out-of-distribution (OOD) detection as the task of evaluating whether a sample comes from the in-distribution on which a neural network was trained. They also introduced benchmark datasets and metrics for this task. Additionally, Hendrycks and Gimpel established the baseline performance by proposing an OOD detection approach that uses the maximum predicted probability (MPS) as the score to detect OOD examples. 

OOD detection is closely related to anomaly detection \cite{10.1145/3439950}. {\color{black}However, in OOD detection, we have multiple (usually many more than two) classes.}
From an anomaly detection perspective, examples that belong to any of these classes are considered ``normal''. Additionally, we have labels that individually identify examples from each of these ``normal'' classes, which collectively represent what we call the in-distribution. There are no training examples of the ``abnormal'' class, which are called out-of-distribution examples in the context of {\color{black}OOD} detection. We have to decide whether we have an in-distribution (``normal'') example or an out-of-distribution (``abnormal'') example during inference. In the first case, we additionally have to predict the correct class. 

\begin{figure}%[t]
\centering
\vskip -0.2cm
\includegraphics[width=0.45\textwidth,trim={0 0 0 0},clip]{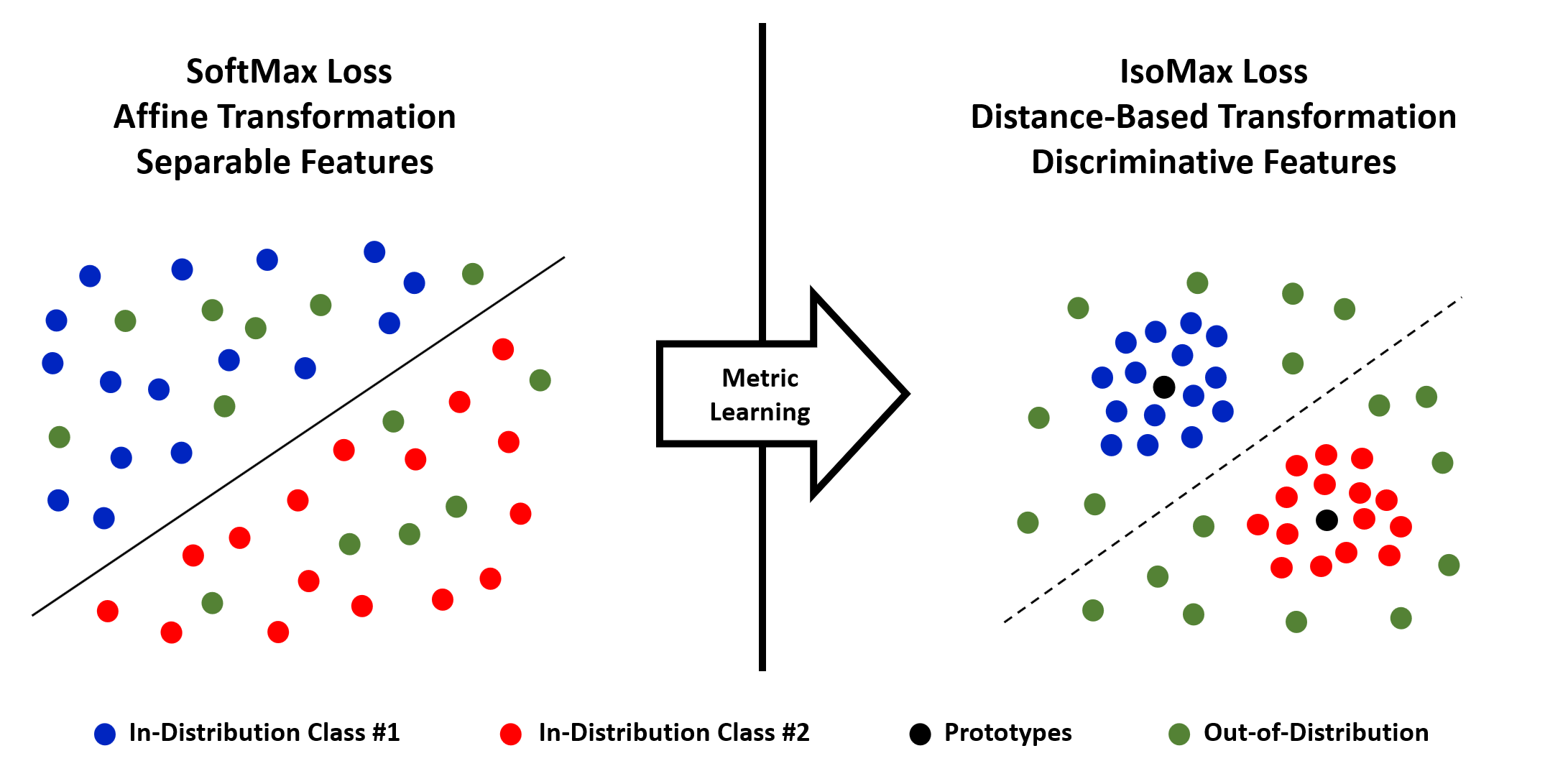} 
\vskip -0.2cm
\caption{Adapted from \cite[Fig.~1]{DBLP:conf/eccv/WenZL016}. SoftMax loss produces separable features. Metric learning on features extracted from SoftMax loss trained networks may convert from the situation on the left to the situation on the right. Exclusively distance-based (isotropic) losses tend to generate more discriminative features. Out-of-distribution examples are more discernible when in-distribution examples are concentrated around prototypes.}
\label{fig:separable-discriminative}
\end{figure}

We argue that the unsatisfactory OOD detection performance of modern neural networks is mainly due to the drawbacks of the currently used loss functions rather than model~limitations. The SoftMax loss\footnote{We follow the ``SoftMax loss'' expression as defined in \cite{liu2016large}.}  %(Fig.~\ref{fig:softmax_loss})
anisotropy does not incentivize the concentration of high-level representations in the feature space~\cite{DBLP:conf/eccv/WenZL016, Hein2018WhyRN}, which makes OOD detection difficult~\cite{Hein2018WhyRN}. Indeed, its affine transformation produces separable rather than discriminative features, which would limit OOD detection (Fig.~\ref{fig:separable-discriminative}). Additionally, SoftMax loss produces overconfident predictions \cite{Guo2017OnCO}, equivalent to low mean entropy probability distributions, which is in frontal disagreement with the maximum entropy principle \cite{PhysRev.106.620, PhysRev.108.171, 10.5555/1146355}.

Rather than directly fixing the above-mentioned critical SoftMax loss drawbacks, current OOD detection approaches improve OOD detection performance by adding novel techniques to circumvent its weakness. However, this strategy usually produces undesired side effects and adds problematic requirements to the solution (Table~\ref{tab:overview}). For example, Out-of-DIstribution detector for Neural networks (ODIN) \cite{liang2018enhancing} and the Mahalanobis distance-based method \cite{lee2018simple} use \emph{input preprocessing}, which produces remarkably slow and energy-inefficient inferences~\cite{thompson2020computational}. Additionally, to tune hyperparameters, these solutions need unrealistic access to OOD samples or adversarial examples, which require a cumbersome process to generate.

In other situations, similar to the Mahalanobis distance-based approach, \cite{Scheirer_2013_TPAMI,Scheirer_2014_TPAMIb,7298799,Rudd_2018_TPAMI} require metric learning on features extracted from pretrained~models. Another drawback usually present in OOD detection approaches is the \emph{classification accuracy drop} \cite{techapanurak2019hyperparameterfree, Hsu2020GeneralizedOD}, which is a harmful side effect because classification is commonly the primary aim of the system \cite{carlini2019evaluating}. Methods based on adversarial training, such as Adversarial Confidence Enhancing Training (ACET) \cite{Hein2018WhyRN}, often increase training times \cite{DBLP:conf/iclr/WongRK20} and present limited scalability when dealing with large-size images \cite{DBLP:conf/nips/ShafahiNG0DSDTG19}. Furthermore, adversarial training may produce classification accuracy~drop \cite{Raghunathan2019AdversarialTC}.

In some cases, OOD detection proposals require architecture modifications \cite{yu2019unsupervised} or ensemble methods \cite{vyas2018out, lakshminarayanan2017simple}. Despite significantly improving the OOD detection performance, loss enhancement (regularization) techniques, such as outlier exposure \cite{hendrycks2018deep, papadopoulos2019outlier}, background methods \cite{NIPS2018_8129}, and the energy-based fine-tuning \cite{DBLP:journals/corr/abs-2010-03759} require the addition of carefully chosen extra/outlier/background data and expand memory usage. Moreover, they usually add hyperparameters to the solution.

Solutions based on uncertainty (or confidence) estimation (or calibration) \cite{kendall2017uncertainties,Leibig2017LeveragingUI,malinin2018predictive,kuleshov2018accurate,subramanya2017confidence} usually present additional complexity, slow and energy-inefficient inferences \cite{thompson2020computational}, and OOD detection performance worse than ODIN~\cite{shafaei2018biased,Hsu2020GeneralizedOD}.

\emph{In this paper, rather than circumvent SoftMax loss limitations, we follow a different strategy: we propose replacing it altogether with a novel loss that does not present the SoftMax loss mentioned drawbacks}. Additionally, we used entropy for OOD detection. By doing so, we were able to improve the neural networks' OOD detection baseline performance significantly. \emph{Using this novel baseline OOD detection approach, previously mentioned or future techniques may improve the neural networks' OOD detection performance even further}. Therefore, we propose IsoMax, a loss that is isotropic (exclusively distance-based) and produces high entropy posterior probability distributions in agreement with the maximum entropy principle. We also propose using the entropy of the output probabilities as a score to perform OOD detection.  

As we aim to increase the neural networks' baseline OOD detection performance, we do not rely on collecting additional/outlier/background data and validating the appropriated hyperparameters to improve the OOD detection results \cite{hendrycks2018deep, papadopoulos2019outlier,NIPS2018_8129,DBLP:journals/corr/abs-2010-03759}. As mentioned before, these (and other) current OOD techniques may be used in future works to improve our baseline OOD detection performance. Furthermore, with the sole aim to perform OOD detection, we emphasize that, unlike uncertainty/confidence estimation/calibration methods, our intention is not to produce calibrated probabilities. In the opposite direction, we plan to force the network to provide a posterior probability distribution with the highest possible entropies (as stated by the maximum entropy principle), which correspond to low maximum probabilities.

Networks trained using IsoMax loss produce accurate predictions, as no classification accuracy drop is observed compared to networks trained with SoftMax loss. Additionally, the models trained using IsoMax loss provide \emph{inferences that are fast and have energy and computational efficiency equivalent to models trained with SoftMax loss}, making our solution viable from an economical and environmental point of view~\cite{Schwartz2019GreenA}. \textcolor{black}{Moreover, our approach does not require validation using unrealistic access to OOD samples or the generation of adversarial examples.} Furthermore, our solution is turnkey since it does not require additional/outlier data, feature extraction, metric learning, or hyperparameter tuning. No extra procedures other than typical network training are required.% Considering our approach avoids ad hoc techniques and associated troublesome requirements and undesired side effects, we say that IsoMax loss works as a SoftMax loss drop-in replacement and that the overall solution is seamless. 

We provide insights based on the maximum entropy principle to explain why our approach works. Experimental evidence confirms our theoretical assumptions and shows that our straightforward OOD detection solution significantly outperforms the SoftMax loss performance and is even competitive with OOD detection methods without requiring their supplementary techniques and associated side effects (Table~\ref{tab:overview}).

Indeed, despite being proposed as a baseline OOD detection solution that avoids current approaches drawbacks (e.g., classification accuracy drop, slow and energy-inefficient inferences, hyperparameter tuning, feature extraction, metric learning, and additional data), our approach provides high OOD detection performance. \emph{Naturally, additional performance gains may be obtained in the future by adapting OOD methods to use IsoMax loss rather than SoftMax loss}.

In summary, the major contributions of this article are:

\begin{enumerate}
\item The intuitions that associate the unsatisfactory OOD detection performance of current neural networks with the SoftMax loss anisotropy and disagreement with the maximum entropy principle.
%\newpage
\item The IsoMax loss, which is isotropic, in agreement with the maximum entropy principle, and works as a \emph{seamless SoftMax loss drop-in replacement}. The IsoMax loss trained models present accurate predictions (no classification accuracy drop) and fast inferences that are energy- and computation-efficient. \textcolor{black}{Besides, it does not require additional/outlier/background data. Under these severe restrictions, our solution provides \emph{state-of-the-art} OOD detection performance.}
\item The experimental demonstration that entropy, which is a fast and energy-efficient computation, provides high OOD detection performance when the posterior probability distribution follows the maximum entropy principle, i.e., it presents high mean entropy. When using the entropy, rather than just one output, all network outputs are considered.
\item Our approach improves deep neural networks' baseline OOD detection performance. In applications where the previously mentioned requirements and side effects of current OOD techniques are not a concern, future works may combine them with our loss to achieve even higher OOD detection performance.
%\textcolor{blue}{\item The experimentation that shows that, despite being proposed as a baseline solution that avoids current approaches drawbacks such as classification accuracy drop, slow and energy-inefficient inferences, hyperparameter tuning, feature extraction, metric learning, and additional data, our baseline solution provides near state-of-the-art OOD detection performance.}
\end{enumerate}

%We organize this article as follows.
In Section~\ref{sec:background}, we emphasize the technical aspects of modern OOD detection approaches. In Section~\ref{sec:seamless_approach}, we present our proposal for a novel baseline OOD detection method. We present the experiments, results, and discussions in Section~\ref{sec:experiments}. In Section~\ref{sec:conclusion}, we conclude and present future works. This paper is an expanded version of \cite{macdo2019isotropic}.

\section{Background}\label{sec:background}

\subsection{Distance-based Losses}
%\label{apx:distance-based-losses}

%\textcolor{blue}{
Recently, neural network distance-based losses have been proposed in the context of face recognition. For example, the contrastive \cite{DBLP:conf/nips/SunCWT14} and triplet \cite{DBLP:conf/cvpr/SchroffKP15} losses use high-level feature (embeddings) \emph{pairwise} distances. In both cases, the SoftMax \emph{function}\footnote{We follow the ``SoftMax \emph{function}'' expression as defined in \cite{liu2016large}.} is not present, and the \emph{squared} Euclidean distance is used. One of the main drawbacks is the need for using \emph{Siamese neural networks}, which adds complexity to the solution and expands memory requirements during training~\cite{DBLP:conf/eccv/WenZL016}. Additionally, the \emph{triplet sampling} and \emph{pairwise training}, which implicate the recombination of the training samples with dramatic data expansion, slow convergence and instability~\cite{DBLP:conf/eccv/WenZL016}. Finally, no prototypes are learned during training. The challenge to train networks using \emph{purely} distance-based losses while avoiding \emph{triplet sampling} and \emph{pairwise training} was discussed in \cite{DBLP:conf/eccv/WenZL016}, which proposed a \emph{squared} Euclidean distance-based \emph{regularization} method.
%}

%\textcolor{blue}{
The center loss \cite{DBLP:conf/eccv/WenZL016} has two parameters $\alpha$ and $\lambda$. %It was also proposed for \emph{face recognition}.
We call a loss isotropic when its dependency on the high-level features (embeddings) is performed \emph{exclusively} through distances, which are usually calculated from the class prototypes. In this sense, the center loss is \emph{not} isotropic, as it presents an affine transformation in its SoftMax classification term. Therefore, the center loss inherits the previously mentioned drawbacks of the SoftMax loss affine transformation.
%}

%\textcolor{blue}{
In \cite{Snell2017PrototypicalNF}, the authors proposed a solution based on \emph{squared} Euclidean distance to address few-shot learning. However, this approach does not work as a SoftMax loss drop-in replacement, as it does not simultaneously learn high-level features (embeddings) \emph{and prototypes} using \emph{exclusively} stochastic gradient descent (SGD) and \emph{end-to-end} backpropagation. Indeed, despite learning embeddings using regular SGD and backpropagation, \emph{additional offline procedures are required to calculate the class prototypes} in the mentioned approach.
%}

\subsection{Out-of-Distribution Detection}

ODIN was proposed in \cite{liang2018enhancing} by combining input preprocessing with \emph{temperature calibration}, which consists of changing the scale of the logits of a pretrained model. Despite significantly outperforming the SoftMax loss, the input preprocessing introduced in ODIN considerably increases the inference time by requiring an initial forward pass, a backpropagation, and finally a second forward pass to perform an inference that can be used for OOD detection. Considering that backpropagation is typically slower than a forward pass, \emph{input preprocessing makes ODIN inferences at least three times slower than normal}. Additionally, input preprocessing multiplies the inference power consumption and computational cost by a factor of at least three. Those are severe limitations from an economic and environmental perspective \cite{Schwartz2019GreenA}\footnote{\url{https://www.youtube.com/watch?v=KnOpWgUCtaM}}. Several subsequent OOD detection proposals incorporated input preprocessing and its drawbacks \cite{liang2018enhancing,lee2018simple,Hsu2020GeneralizedOD,DeVries2018LearningNetworks}. Both input preprocessing and temperature calibration \emph{require hyperparameter tuning}.

Consequently, ODIN requires unrealistic access to OOD samples to validate hyperparameters. Even if the supposed OOD samples are available during design-time, using these examples to tune hyperparameters makes the solution overfit to detect this particular type of out-distribution\footnote{In this paper, the expression ``out-distribution'' refers to any distribution that generates OOD samples.}. The system will likely operate under unknown out-distributions in real-world applications, and the estimated OOD detection performance could degrade significantly. Therefore, using OOD samples to validate hyperparameters may produce over-optimistic OOD detection performance~estimations \cite{shafaei2018biased}.

A method to avoid using OOD samples was proposed in \cite{Hsu2020GeneralizedOD}. The mentioned method uses only \emph{in-distribution} validation data for tuning the required hyperparameters. However, the proposed loss produced a significant classification accuracy drop in some situations. Moreover, the \emph{in-distribution} CIFAR10/100 validation sets were used for both hyperparameter tuning and OOD detection evaluation. Therefore, the classification accuracy drop reported in \cite{Hsu2020GeneralizedOD} is probably underestimated. In practice, the necessary \emph{in-distribution} validation data for hyperparameter tuning will have to be removed from the training data, which will presumably decrease even further the classification accuracy and, consequently, also the OOD detection performance. Additionally, \cite{Hsu2020GeneralizedOD} also uses input preprocessing, making inferences considerably inefficient from an environment, economic, and energy perspective.

The Mahalanobis distance-based method\footnote{For the rest of this paper, the expression ``the Mahalanobis distance-based method'' is replaced by ``the Mahalanobis method''.} \cite{lee2018simple} overcomes the need for access to OOD samples by validating hyperparameters using adversarial examples and producing more realistic OOD detection performance estimates. %Hence, in this work, we only consider validation on adversarial samples for competing methods.
However, using adversarial examples has the disadvantage of adding a cumbersome procedure to the solution. Even worse, the generation of adversarial samples itself requires hyperparameter tuning such as the maximum perturbations. While adequate hyperparameters may be known for research datasets, they may be challenging to find for novel real-world~data.

Moreover, as the Mahalanobis approach also requires input preprocessing, the previously mentioned drawbacks associated with this technique are still present in the Mahalanobis solution. \emph{Feature ensemble} introduced in this approach also presents limitations. Indeed, \emph{feature ensembles} require training of extra classification and regression models on features extracted from many network layers. For applications using real-world large-size images, these shallow models may present scalability problems, as they would be required to work in spaces of tens of thousands of dimensions. Finally, the Mahalanobis method is not turnkey, as it involves feature extraction and metric learning post-processing.

Hyperparameter tuning is also a drawback to methods based on adversarial training (e.g., ACET \cite{Hein2018WhyRN}), as we need adequate adversarial perturbations. Additionally, adversarial training is known for increasing training time \cite{DBLP:conf/iclr/WongRK20} and reducing classification accuracy \cite{Raghunathan2019AdversarialTC}. Furthermore, solutions based on adversarial training may not scale to applications dealing with real-world large-sized~images~\cite{DBLP:conf/nips/ShafahiNG0DSDTG19}, which may also be the case of approaches with high computational cost inferences.

The Entropic Open-Set loss and the Objectosphere loss were proposed in \cite{NIPS2018_8129}. These two losses used background samples to improve the performance of detecting unknown inputs. The Entropic Open-Set loss works like the usual SoftMax loss in the in-distribution training data, producing low entropy for these samples. However, it forces maximum entropy in the background samples. {\color{black}The Objectosphere loss is the Entropic Open-Set loss with an added regularization factor that forces} the feature magnitude of in-distribution samples to be near a predefined value $\xi$ while minimizing the feature magnitude of background samples.

\textcolor{black}{Domain-specialized methods have also been proposed. Tack et al. \cite{DBLP:conf/nips/TackMJS20} used data augmentation for image data to improve OOD detection in a self-supervised setting. Sastry et al. \cite{Sastry2019DetectingOE} analyzed statistics of the activations of the pretrained model on training and validation data to detect OOD examples. Hence, we believe one of the main advantages of our approach is that it is \emph{domain-agnostic}, as it can be applied to any domain.}  

\section{Entropic Out-of-Distribution Detection}\label{sec:seamless_approach}

\subsection{Isotropy Maximization Loss}

To mitigate the SoftMax loss drawbacks (i.e., anisotropy and low entropy posterior probability distributions), we design the IsoMax loss by imposing isotropy combined with high (almost maximum) entropy posterior probability distributions, which is obtained using what we call ``the entropy maximization trick''.

\subsubsection{Isotropy}\label{isotropy}

To fix the SoftMax loss anisotropy caused by its affine transformation, we propose the IsoMax loss, which is isotropic in the sense that its logits depend \emph{exclusively} on distances from high-level features to class prototypes. For more information regarding the SoftMax loss anisotropy, how it harms the OOD detection performance, and the advantages of circumventing this problem using an isotropic loss, please see Supplementary Material \ref{apx:softmax-loss-anisotropy}.

We designed the IsoMax loss to work as a SoftMax loss drop-in replacement. Therefore, the swap of the SoftMax loss with the IsoMax loss %(Fig.~\ref{fig:isomax_loss})
requires changes in neither the model's architecture nor training procedures or parameters.

Let $\bm{f}_{\bm{\theta}}(\bm{x})$ represent the high-level feature (embedding) associated with $\bm{x}$, $\bm{p}_{\bm{\phi}}^j$ represent the prototype associated with the class $j$, and $d()$ represent a distance. To construct a isotropic loss, we need to avoid \emph{direct} dependency on $\bm{f}_{\bm{\theta}}(\bm{x})$ or $\bm{p}_{\bm{\phi}}^j$. Therefore, the loss has to be a function that \emph{exclusively} depends on the embedding-prototype distances given by $d({\bm{f}_{\bm{\theta}}(\bm{x}),\bm{p}_{\bm{\phi}}^j})$. Therefore, we can write:

\begin{align}\label{eq:isotropic_loss1}
\begin{split}
\mathcal{L}_{I}\!=\!g(d({\bm{f}_{\bm{\theta}}(\bm{x}),\bm{p}_{\bm{\phi}}^j}))
\end{split}
%\end{equation}
\end{align}

In the previous equation, $g()$ represents a scalar function. The expression $d({\bm{f}_{\bm{\theta}}(\bm{x}),\bm{p}_{\bm{\phi}}^j})$ represents the isotropic layer, where its weights are given by the learnable prototypes $\bm{p}_{\bm{\phi}}^j$.

We decided to normalize the embedding-prototype distances using the SoftMax \emph{function} to allow interpretation in terms of probabilities. Therefore, the embedding-prototype distances represent the logits of the SoftMax \emph{function} and correspond to the output of the isotropic layer. We also decided to use cross-entropy for efficient optimization. Hence, we can write the following equation:

%\begin{multline}
\begin{align}\label{eq:isotropic_loss2}
\begin{split}
\mathcal{L}_{I}(\hat{y}^{(k)}|\bm{x})
=-\log\left(\frac{\exp(-d(\bm{f}_{\bm{\theta}}(\bm{x}),\bm{p}_{\bm{\phi}}^k))}{\sum\limits_j\exp(-d(\bm{f}_{\bm{\theta}}(\bm{x}),\bm{p}_{\bm{\phi}}^j))}\right)
%\end{multline}
\end{split}
\end{align}

In the above equation, $\hat{y}^{(k)}$ represents the label of the correct class, while the negative logarithm represents the cross-entropy. The negative terms before the distances are necessary to indicate the negative correlation between distances and probabilities. The expression between the outermost parentheses applied to the term $-d({\bm{f}_{\bm{\theta}}(\bm{x}),\bm{p}_{\bm{\phi}}^j})$ represents the SoftMax \emph{function}. We use the non-squared Euclidean distance for $d(.,\!.)$, and justify its use in Supplementary Material \ref{apx:euclidean-distance}.

Unlike usual metric learning-based OOD detection approaches, rather than learning a metric from a preexisting feature space (metric learning on features extracted from a pretrained model), when using the IsoMax loss, we learn a feature space that is, from the start, consistent with the chosen metric. Indeed, the minimization of Equation~\eqref{eq:isotropic_loss2} is achieved by making the expression inside the outer parentheses goes to one. This is only possible by reducing the distances between the high-level features (embeddings) and the associated class prototypes while simultaneously keeping high distances among class prototypes. Hence, the main aim of metric learning, which is to reduce intraclass distances while increasing interclass distances, is performed naturally during the neural network training, avoiding feature extraction and metric learning post-processing phases.

% There is no need to map the features produced by the network to an ad hoc embedding space or to learn a distance since the network high-level features are learned from the start to be consistent with the Euclidean distance.% Additionally, \emph{prototypes are not learned seamlessly during the network backpropagation training}. In IsoMax, \emph{to build a SoftMax loss drop-in replacement with improved OOD detection performance, the prototypes are treated as usual parameters learned during the regular end-to-end backpropagation}.

%While \cite{6517188} used the Mahalanobis distance, \cite{Snell2017PrototypicalNF} proposed the \emph{squared} Euclidean distance. However, the covariance matrix makes hard using the Mahalanobis distance to directly train a neural network.
%Finally, we experimentally observed that using \emph{nonsquared} Euclidean distance performed better than \emph{squared} Euclidean distance.

\subsubsection{Entropy Maximization}\label{entropy_maximization}

Isotropy increases OOD detection performance. However, we need to circumvent the SoftMax loss propensity to produce significantly low entropy posterior probability distributions for further gains. To achieve high entropy probability distributions in agreement with the maximum entropy principle, we introduce the \emph{entropic scale}, a constant scalar multiplicative factor applied to the logits \emph{throughout training} that is, nevertheless, \emph{removed before inference}. We call this procedure ``the entropy maximization trick''.

Our experiments show that the proposed strategy forces the neural networks to produce outputs with significantly high entropy probability distributions (as recommended by the maximum entropy principle) rather than the usual low entropy (overconfident) posterior probability distributions.

The entropic scale is related to the \emph{inverse of the temperature} of the SoftMax \emph{function}. However, training with a \emph{predefined constant} entropic scale and then \emph{removing it before inference} is \emph{completely} different from temperature calibration. On the one hand, in ODIN and similar methods based on temperature calibration, the temperature of a \emph{pretrained} model is validated \emph{after training}, which nevertheless was performed with a temperature equal to one. This validation usually requires unrealistic access to OOD or adversarial examples, which additionally make them not turnkey. Additionally, over-optimistic performance estimation is commonly produced~\cite{shafaei2018biased}. On the other hand, our approach requires neither hyperparameter validation nor access to OOD or adversarial data.

The principle of maximum entropy, formulated by E. T. Jaynes to unify the statistical mechanics and information theory entropy concepts \cite{PhysRev.106.620, PhysRev.108.171}, states that when estimating probability distributions, we should choose the one that produces the maximum entropy consistent with the given constraints \cite{10.5555/1146355}. Following this principle, we avoid introducing additional assumptions or bias not presented in the data. 
%\footnote{\url{https://mtlsites.mit.edu/Courses/6.050/2003/notes}

Hence, from a set of trial probability distributions that satisfactorily describe the prior knowledge available, the distribution that presents the maximal information entropy (i.e., the least informative option) represents the best possible choice. In other words, we must produce posterior probability distributions as underconfident as possible as long as they match the correct predictions. 

The principle of maximum entropy has been studied as a \emph{regularization factor} \cite{DBLP:conf/nips/DubeyGRN18,pereyra2017regularizing}. In some cases, it has also been used as a direct optimization procedure without connection to cross-entropy minimization or backpropagation. For example, in \cite{514879,berger-etal-1996-maximum,pmlr-v5-shawe-taylor09a}, the maximization of the entropy subject to a constraint on the expected classification error was shown to be equivalent to solving an unconstrained Lagrangian.

The relation between the principle of maximum entropy and Bayesian methods has been studied in \cite{Jaynes1988}. Despite being theoretical well-grounded \cite{10.5555/534975,Williamson2005ObjectiveBN,Pages493-533,DBLP:journals/mima/Hosni13}, \emph{direct} entropy maximization presents high computational complexity, as it is an NP-complete problem \cite{10.5555/534975,Pages493-533}. Alternatively, modern neural networks are trained using computationally efficient cross-entropy. However, this procedure does not prioritize high entropy (low confidence) posterior probability distributions. \emph{Actually, the opposite is true}. Indeed, the minimization of cross-entropy has the undesired side effect of producing low entropy (overconfident) posterior probability distributions~\cite{Guo2017OnCO}.

Unlike the previously mentioned works, we use the principle of maximum entropy neither to motivate the construction of regularization mechanisms, such as label smoothing or the confidence penalty \cite{DBLP:conf/nips/DubeyGRN18,pereyra2017regularizing}, nor to perform \emph{direct} maximum entropy optimization \cite{514879,berger-etal-1996-maximum,pmlr-v5-shawe-taylor09a}. The entropy is not even calculated during IsoMax loss training. In the opposite direction, we use the principle of maximum entropy as a theoretical motivation for constructing high entropy (low confidence) posterior probabilities, still relying on computationally efficient cross-entropy minimization. Since our approach does \emph{not} directly maximize the entropy, we cannot state that IsoMax produces the maximum entropy posterior probability distribution. However, the entropies produced by IsoMax loss are high enough to improve the OOD detection performance significantly.

%\begin{equation}\label{eq:cross_entropy_SoftMax_minimization}
\begin{align}%\label{eq:cross_entropy_softmax_minimization}
%\begin{split}
\mathcal{L}_{\textsf{SoftMax}}\!=\!-\log\left(\frac{\exp(L_k)}{\sum\limits_j\exp(L_j)}\right)\to 0\label{eq:cross_entropy_softmax_minimization1}\\
\implies
\mathcal{P}(y|\bm{x})\to 1\label{eq:cross_entropy_softmax_minimization2}\\
\implies \mathcal{H}_{\textsf{SoftMax}}\to 0\label{eq:cross_entropy_softmax_minimization3}
%\end{split}
%\end{equation}
\end{align}

Equation \eqref{eq:cross_entropy_softmax_minimization1} describes the behavior of the cross-entropy and entropy for the SoftMax loss. $L_j$ represents the logits associated with class $j$, and $L_k$ represents the logits associated with the correct class $k$. When minimizing the loss (Equation \eqref{eq:cross_entropy_softmax_minimization1}), high probabilities are generated (Equation \eqref{eq:cross_entropy_softmax_minimization2}). Consequently, significantly low entropy posterior probability distributions are produced (Equation \eqref{eq:cross_entropy_softmax_minimization3}). Hence, the usual cross-entropy minimization tends to generate unrealistic overconfident (low entropy) probability distributions. Therefore, we have an opposition between cross-entropy minimization and the principle of maximum~entropy.

%\begin{equation}\label{eq:cross_entropy_isomax_minimization}
\begin{align}%\label{eq:cross_entropy_isomax_minimization}
%\begin{split}
\mathcal{L}_{\textsf{IsoMax}}\!=\!-\log\left(\frac{\exp(-E_s\!\times\!D_k)}{\sum\limits_j\exp(-E_s\!\times\!D_j)}\right)\to 0\label{eq:cross_entropy_isomax_minimization1}\\
\centernot\implies \mathcal{P}(y|\bm{x})\to 1\label{eq:cross_entropy_isomax_minimization2}\\
\centernot\implies \mathcal{H}_{\textsf{IsoMax}} \to 0\label{eq:cross_entropy_isomax_minimization3}
%\end{split}
%\end{equation}
\end{align}

The IsoMax loss conciliates these contradictory objectives (loss minimization and entropy maximization) by multiplying the logits by a constant positive scalar $E_s$, which is presented during training but removed before inference. Equation \eqref{eq:cross_entropy_isomax_minimization1} demonstrates how the entropic scale (presented at training time but removed at inference time) allows the production of high entropy posterior distributions despite using cross-entropy minimization.  $D_j$ represents the distances associated with class $j$, and $D_k$ represents the distances associated with the correct class $k$. The $E_s$ present during training allows the term $-E_s\!\times\!D_k$ to become high enough (less negative compared to $-E_s\!\times\!D_j$) to produce low loss (Equation \eqref{eq:cross_entropy_isomax_minimization1}) \emph{without} producing high probabilities for the correct classes, as they are calculated with the $E_s$ removed (Equation \eqref{eq:cross_entropy_isomax_minimization2}). Thus, it is possible to build posterior probability distributions with high entropy (Equation \eqref{eq:cross_entropy_isomax_minimization3}) in agreement with the fundamental principle of maximum entropy despite using cross-entropy to minimize the loss.

We emphasize that, regardless of training with the entropic scale, if we do not remove it before performing inference, the IsoMax loss produces outputs with entropies as low as those produced by the SoftMax loss, the OOD detection performance does not further increase. Hence, returning to Equation \eqref{eq:isotropic_loss2}, multiplying the embedding-prototype distances by $E_s$, and making $d()$ equal to the \emph{nonsquared} Euclidean distance, we write the definitive IsoMax loss as:

%\begin{multline}
\begin{align}\label{eq:loss_isomax}
\begin{split}
\mathcal{L}_{I}(\hat{y}^{(k)}|\bm{x})
=-\log^\dagger\left(\frac{\exp(-E_s\norm{\bm{f}_{\bm{\theta}}(\bm{x})\!-\!\bm{p}_{\bm{\phi}}^k})}{\sum\limits_j\exp(-E_s\norm{\bm{f}_{\bm{\theta}}(\bm{x})\!-\!\bm{p}_{\bm{\phi}}^j})}\right)
\\
=-\log^\dagger\left(\frac{\exp(-E_s\sqrt{(\bm{f}_{\bm{\theta}}(\bm{x})\!-\!\bm{p}_{\bm{\phi}}^k)\!\cdot\!(\bm{f}_{\bm{\theta}}(\bm{x})\!-\!\bm{p}_{\bm{\phi}}^k)})}{\sum\limits_j\exp(-E_s\sqrt{(\bm{f}_{\bm{\theta}}(\bm{x})\!-\!\bm{p}_{\bm{\phi}}^j)\!\cdot\!(\bm{f}_{\bm{\theta}}(\bm{x})\!-\!\bm{p}_{\bm{\phi}}^j)})}\right)
%\end{multline}
\end{split}
\end{align}
\blfootnote{\textsuperscript{$\dagger$}\emph{The probability (i.e., the expression between the outermost parenteses) and logarithm operations are computed sequentially and separately for higher OOD detection performance (please, see the source code).}}

We emphasize that removing the entropic scale after the training does not affect the solution's ability to represent the prior knowledge available, as it does not change the predictions. Therefore, the expression for the probabilities with the entropic scale removed is preferable, as it increases the entropy of the posterior distribution in agreement with the principle of maximum entropy. %Therefore, unlike SoftMax loss, IsoMax loss is in accord with the principle of maximum entropy.
Hence, the inference probabilities for the IsoMax loss are defined as follows:

%\begin{multline}
\begin{align}\label{eq:probability_isomax}
\begin{split}
p_{I}(y^{(i)}|\bm{x})
&=\frac{\exp(-\norm{\bm{f}_{\bm{\theta}}(\bm{x})\!-\!\bm{p}_{\bm{\phi}}^i})}{\sum\limits_j\exp(-\norm{\bm{f}_{\bm{\theta}}(\bm{x})\!-\!\bm{p}_{\bm{\phi}}^j})}\\
%&=\frac{\exp(-\beta\sqrt{(\bm{f}_{\bm{\theta}}(\bm{x})\!-\!\bm{p}_{\bm{\phi}}^k)\cdot(\bm{f}_{\bm{\theta}}(\bm{x})\!-\!\bm{p}_{\bm{\phi}}^k)})}{\sum\limits_{j}\exp(-\beta\sqrt{(\bm{f}_{\bm{\theta}}(\bm{x})\!-\!\bm{p}_{\bm{\phi}}^j)\cdot(\bm{f}_{\bm{\theta}}(\bm{x})\!-\!\bm{p}_{\bm{\phi}}^j)})}\\
&=\frac{\exp(-\sqrt{(\bm{f}_{\bm{\theta}}(\bm{x})\!-\!\bm{p}_{\bm{\phi}}^k)\!\cdot\!(\bm{f}_{\bm{\theta}}(\bm{x})\!-\!\bm{p}_{\bm{\phi}}^k)})}{\sum\limits_j\exp(-\sqrt{(\bm{f}_{\bm{\theta}}(\bm{x})\!-\!\bm{p}_{\bm{\phi}}^j)\!\cdot\!(\bm{f}_{\bm{\theta}}(\bm{x})\!-\!\bm{p}_{\bm{\phi}}^j)})}
%\end{multline}
\end{split}
\end{align}

Please, notice that the entropic scale $E_s$ was intentionally removed from the loss Equation \ref{eq:loss_isomax} to build the inference Equation \ref{eq:probability_isomax} to perform ``the entropy maximization trick''.  

\subsubsection{Initialization and Implementation Details}

We experimentally observed that using the Xavier \cite{glorot2010understanding} or Kaiming \cite{He2016DelvingClassification} initializations for prototypes makes the OOD detection performance oscillate. Sometimes it improves, sometimes it decreases. Hence, we decided to initialize all prototypes to zero vector, as this is the most natural value for untrained embeddings. Weight decay is applied to the prototypes, as they are regular trainable parameters.

To calculate losses based on cross-entropy, deep learning libraries usually combine the probability and logarithm calculations into a single computation. \emph{However, we experimentally observed that sequentially computing these calculations as stand-alone operations improves IsoMax performance}.

The class prototypes have the same dimension as the neural network last layer representations. Naturally, the number of prototypes is equal to the number of classes. Therefore, the IsoMax loss has fewer parameters than the SoftMax loss, as it has no bias to be learned.

Finally, we verified classification accuracy drop and low or oscillating OOD detection performance when trying to integrate $E_s$ with cosine similarity \cite{liu2016large, DBLP:journals/spl/WangCLL18, DBLP:conf/cvpr/DengGXZ19} or the affine transformations used in SoftMax loss. In such cases, the above trick to initialize the prototypes to zero vector cannot~be~performed. Moreover, the Mahalanobis distance cannot be used directly during the neural network training because the covariance matrix is not differentiable. Therefore, we confirmed \emph{nonsquared} Euclidean distance as the best option to build the IsoMax~loss. 

\begin{figure*}%[t]
%\vskip -0.25 cm
\centering
\subfloat[]{\includegraphics[width=0.2375\textwidth]{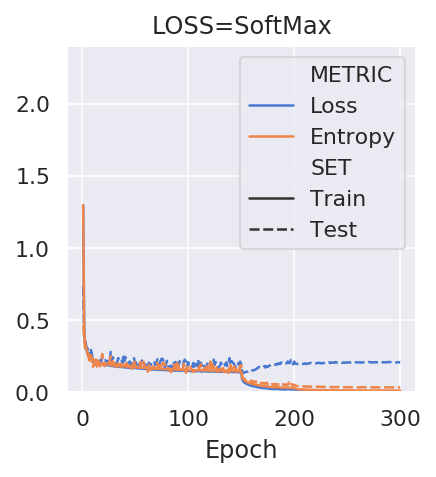}\label{fig:train_loss_entropies_softmax}}
\subfloat[]{\includegraphics[width=0.2375\textwidth]{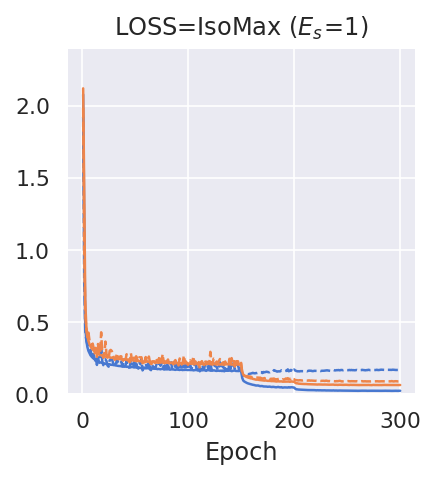}\label{fig:train_loss_entropies_isomax1}}
\subfloat[]{\includegraphics[width=0.2375\textwidth]{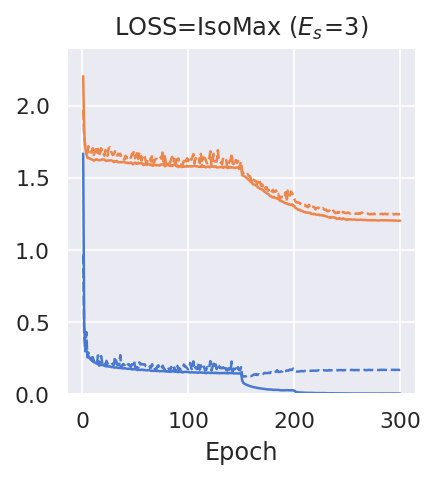}\label{fig:train_loss_entropies_isomax3}}
\subfloat[]{\includegraphics[width=0.2375\textwidth]{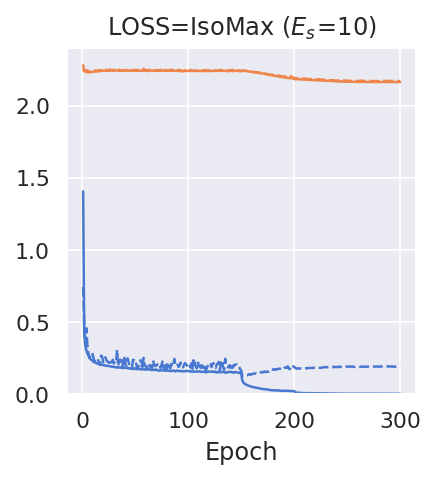}\label{fig:train_loss_entropies_isomax10}}
\\
%\vskip 0.05cm
%\vskip -0.05cm
\subfloat[]{\includegraphics[width=0.925\textwidth]{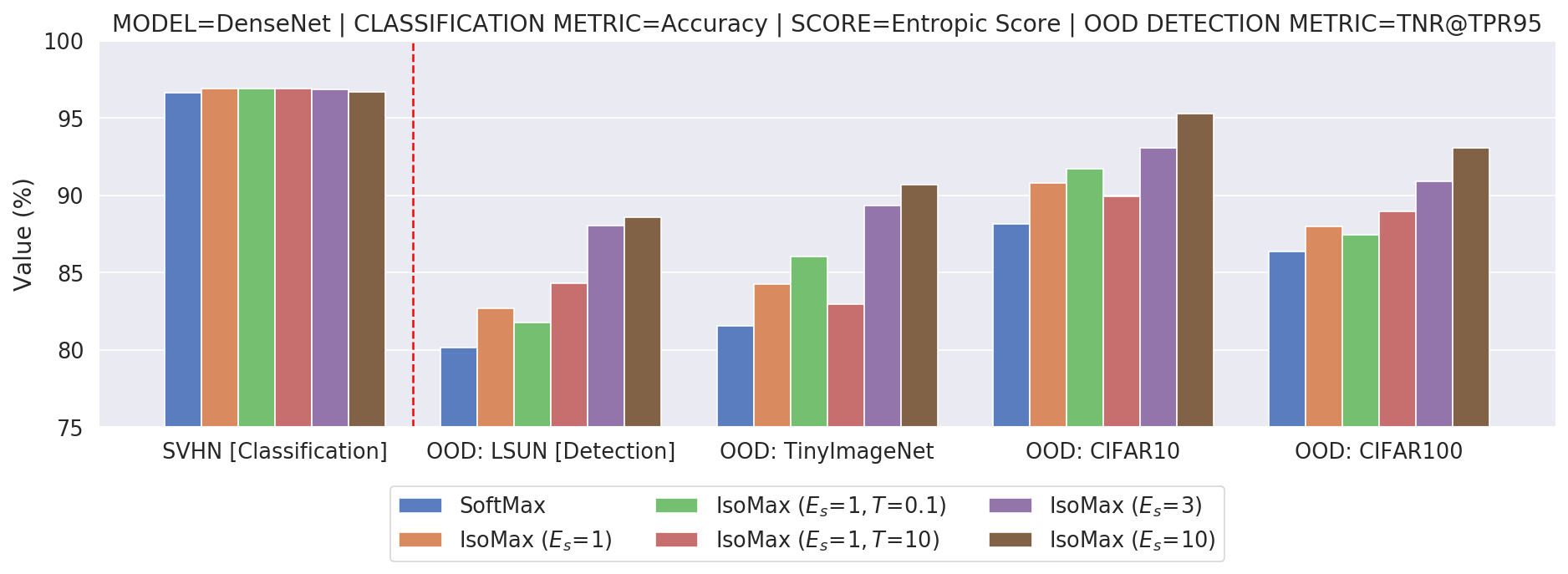}\label{fig:entropic_scale_parametrization}}
\\
\subfloat[]{\includegraphics[width=0.925\textwidth]{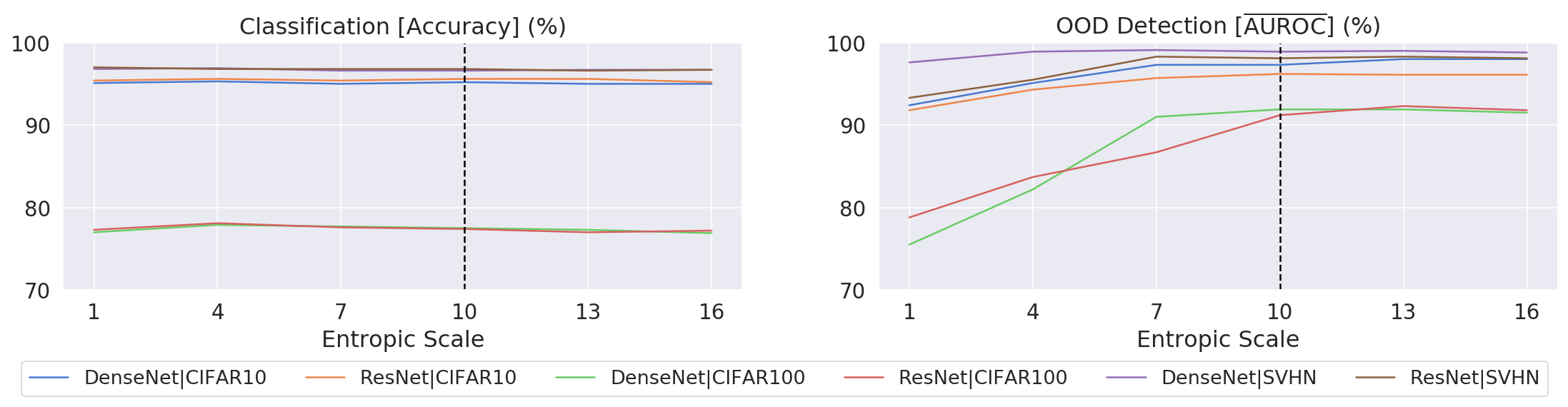}\label{fig:entropic_score_study}}
%\vskip -0.25cm
%\vskip -0.05cm
\caption{IsoMax loss effects: (a) SoftMax loss simultaneously minimizes both the cross-entropy and the entropy of the posterior probabilities. (b)~IsoMax loss produces low entropy posterior probabilities for a low entropic scale \mbox{($E_s\!\!=\!1$)}. (c) IsoMax loss produces medium mean entropy for an intermediate entropic scale \mbox{($E_s\!\!=\!3$)}. (d) IsoMax loss minimizes the cross-entropy while producing high mean entropies for the high entropic scale \mbox{($E_s\!\!=\!10$)}. \emph{``The entropy maximization trick'' is the fundamental mechanism that allows us to migrate from low entropy posterior probability distributions (a,b) to high entropy posterior probability distributions (d). Higher entropic scale values correlate to higher mean entropies as recommended by the principle of maximum entropy. Regardless of training with a high entropic scale, if we do not remove it for inference (``the entropy maximization trick''), the IsoMax loss always produces posterior probability distributions with entropies as low as those generated by the SoftMax loss. An entropic scale equal to ten is sufficient to virtually produce the maximum possible entropy posterior probability distribution, as the highest possible value of the entropy is $\log(N)$, where $N$ is the number of classes}. (e) The left side of the dashed vertical red line presents classification accuracies. The dashed vertical red line's right side shows OOD detection performance using the entropic score and the TNR@TPR95 metric. Higher mean entropies produce increased OOD detection performance regardless of the out-distribution. We emphasize that OOD examples were never used during training and no validation was required to tune hyperparameters. Additionally, isotropy enables IsoMax loss to produce higher OOD performance than SoftMax loss, even for the unitary value of the entropic scale. Training using \mbox{$E_s\!\!=\!1$} and then making the temperature \mbox{$T\!\!=\!0.1$} or \mbox{$T\!\!=\!10$} during inference produces lower {\color{black}OOD} detection performance than training using \mbox{$E_s\!\!=\!10$} and removing it for inference, which consists in our proposal. Finally, IsoMax loss presents similar classification accuracy compared with SoftMax regardless of the entropic scale. The entropic score, which is the negative entropy of the network output probabilities, was used as the score. (f) Study of classification accuracy and OOD detection performance dependence on the entropic scale. $\overline{\mathrm{AUROC}}$ represents the mean AUROC considering all out-of-distribution data. The classification accuracy and the mean OOD detection performance are approximately stable for \mbox{$E_s\!\!=\!10$} or higher regardless of the dataset and model. It also does not depend on the number of training classes N, which is probably explained by the fact that the entropic scale is inside an exponential function while the entropy increases only with the logarithm of the number of classes. Making $E_s$ learnable did not considerably improve or decrease the OOD detection results.}
\label{fig:train_losses_entropies_and_entropic_scale_parametrization}
\end{figure*}

\subsection{Entropic Score}

We define our score to perform OOD detection, called the entropic score, as the output probabilities negative entropy:

%\begin{equation}\label{eq:entropy}
\begin{align}\label{eq:entropic_score}
\mathcal{ES}\!=\!-\sum_{i=1}^{N}{p(y^{(i)}|\bm{x})}\log p(y^{(i)}|\bm{x})
%\end{equation}
\end{align}

Using the negative entropy as a score to evaluate whether a particular sample is OOD, we consider the information provided by all available network outputs rather than just one. For instance, ODIN and ACET only use the maximum probability, while the Mahalanobis method only uses the distance to the nearest prototype. The entropy has been studied before for anomaly detection \cite{DBLP:journals/csur/ChandolaBK09}.

Regarding the use of entropy as an OOD score, one of the most relevant contributions of this paper is to experimentally show that it does not significantly overcome the maximum probability score when dealing with the usual low entropy (overconfident) neural networks' posterior probability distributions. However, the entropic score significantly improves the OOD detection performance when used with the high entropy (underconfident) neural networks' posterior probability distributions produced by the IsoMax loss in agreement with the principle of maximum entropy. For more information regarding the use of entropy as a score to perform OOD, please see Supplementary Material \ref{apx:entropic_score}. For a detailed differentiation between our approach and the proposed in \cite{techapanurak2019hyperparameterfree}, please see Supplementary Material \ref{apx:scaled_cosine}.

\section{Experimental Analyses}\label{sec:experiments}

\begin{table*}%[!t]
\setlength{\tabcolsep}{6pt}
\renewcommand{\arraystretch}{0.5}
%\small
\centering
%\caption{\normalsize Seamless Out-of-Distribution Detection: Fast and Efficient Inferences. Scalable and Turnkey Approaches.}
%\caption{\normalsize Out-of-Distribution Detection: Fast and Energy-Efficient Inferences. Turnkey Approaches.}
\caption{Current and Proposed Baseline Out-of-Distribution Detection Approaches Comparison.\\Fast and Energy-Efficient Inferences. Turnkey Approaches. No extra/outlier/background data used.}
\vspace{-0.05in}
%\begin{tabularx}{\textwidth}{CCC|CCC}
\begin{tabularx}{\textwidth}{lll|CCC}
\toprule
%%%%%%%%%%%%%%%%%%%%%%%%%%%%%%%%%%%%%%%%%%%%%%%%%%%%%%%%%%%%%%%%%%%%%%%%%%%%%%%%%%%%%%%
%\multirow{4}{*}{\begin{tabular}[c]{@{}c@{}}\\Model\end{tabular}} & \multirow{4}{*}{\begin{tabular}[c]{@{}c@{}}\\In-Distribution\\(training)\end{tabular}} & \multirow{4}{*}{\begin{tabular}[c]{@{}c@{}}\\Out-Distribution\\(unseen)\end{tabular}} & 
\multirow{4}{*}{\begin{tabular}[c]{@{}c@{}}\\Model\end{tabular}} & \multirow{4}{*}{\begin{tabular}[c]{@{}c@{}}\\In-Data\\(training)\end{tabular}} & \multirow{4}{*}{\begin{tabular}[c]{@{}c@{}}\\Out-Data\\(unseen)\end{tabular}} & 
%\multirow{4}{*}{\begin{tabular}[c]{@{}c@{}}\\OOD Detection Test Set\\(unseen)\end{tabular}} & 
%%%%%%%%%%%%%%%%%%%%%%%%%%%%%%%%%%%%%%%%%%%%%%%%%%%%%%%%%%%%%%%%%%%%%%%%%%%%%%%%%%%%%%%
%\multicolumn{3}{c}{Seamless Out-of-Distribution Detection:}\\
\multicolumn{3}{c}{Baseline Out-of-Distribution Detection Approaches Comparison.}\\
%&&& \multicolumn{3}{c}{Fast and Efficient Inferences. Scalable and Turnkey Approaches}\\
&&& \multicolumn{3}{c}{Fast and Energy-Efficient Inferences. Turnkey Approaches. No outlier data used.}\\
\cmidrule{4-6}
%&&& \multicolumn{3}{c}{Training Loss + Inference Score}\\
&&& TNR@TPR95 (\%) [$\uparrow$] & AUROC (\%) [$\uparrow$] & DTACC (\%) [$\uparrow$]\\
%&&& TNR@TPR95 (\%) & AUROC (\%) & DTACC (\%)\\
%\cmidrule{4-6}
%&&& \multicolumn{3}{c}{Detection Metric}\\
&&& \multicolumn{3}{c}{SoftMax+MPS \cite{hendrycks2017baseline} / SoftMax+ES / IsoMax+ES (ours)}\\
%%%%%%%%%%%%%%%%%%%%%%%%%%%%%%%%%%%%%%%%%%%%%%%%%%%%%%%%%%%%%%%%%%%%%%%%%%%%%%%%%%%%%%%
\midrule
\multirow{9}{*}{\begin{tabular}[c]{@{}c@{}}DenseNet~~\end{tabular}}
& \multirow{3}{*}{\begin{tabular}[c]{@{}c@{}}CIFAR10~~\end{tabular}} 
& SVHN & 32.1$\pm$0.3 / 33.1$\pm$0.4 / \bf77.1$\pm$0.3 & 86.5$\pm$0.4 / 86.8$\pm$0.3 / \bf96.7$\pm$0.4 & 79.8$\pm$0.3 / 79.8$\pm$0.3 / \bf91.8$\pm$0.3\\
&& TinyImageNet~~& 55.7$\pm$0.3 / 59.7$\pm$0.4 / \bf88.1$\pm$0.4 & 93.5$\pm$0.4 / 94.1$\pm$0.4 / \bf97.9$\pm$0.3 & 87.5$\pm$0.2 / 87.9$\pm$0.3 / \bf93.3$\pm$0.3\\
&& LSUN & 64.8$\pm$0.4 / 69.6$\pm$0.3 / \bf94.6$\pm$0.2 & 95.1$\pm$0.3 / 95.8$\pm$0.2 / \bf98.9$\pm$0.3 & 89.8$\pm$0.3 / 90.1$\pm$0.3 / \bf95.0$\pm$0.4\\
\cmidrule{2-6} 
& \multirow{3}{*}{\begin{tabular}[c]{@{}c@{}}CIFAR100\end{tabular}} 
& SVHN & 20.5$\pm$0.6 / {\bf24.8$\pm$0.8} / {\bf23.6$\pm$0.9} & 80.1$\pm$0.7 / 81.8$\pm$0.7 / \bf88.8$\pm$0.6 & 73.8$\pm$0.6 / 74.4$\pm$0.7 / \bf83.9$\pm$0.6\\
&& TinyImageNet & 19.3$\pm$0.9 / 23.8$\pm$0.8 / \bf49.0$\pm$0.6 & 77.1$\pm$0.6 / 78.7$\pm$0.7 / \bf92.8$\pm$0.6 & 70.5$\pm$0.6 / 71.3$\pm$0.7 / \bf86.5$\pm$0.6\\
&& LSUN & 18.7$\pm$0.6 / 24.3$\pm$0.8 / \bf63.1$\pm$0.6 & 75.8$\pm$0.7 / 77.8$\pm$0.6 / \bf94.8$\pm$0.6 & 69.4$\pm$0.8 / 70.3$\pm$0.9 / \bf89.2$\pm$0.6\\
\cmidrule{2-6} 
& \multirow{3}{*}{\begin{tabular}[c]{@{}c@{}}SVHN\end{tabular}} 
& CIFAR10 & 81.5$\pm$0.2 / 83.7$\pm$0.3 / \bf94.1$\pm$0.2 & 96.5$\pm$0.3 / 96.9$\pm$0.2 / \bf98.5$\pm$0.2 & 91.9$\pm$0.3 / 92.1$\pm$0.2 / \bf95.0$\pm$0.2\\
&& TinyImageNet & 88.3$\pm$0.2 / 90.1$\pm$0.2 / \bf97.2$\pm$0.3 & 97.7$\pm$0.3 / 98.1$\pm$0.2 / \bf99.1$\pm$0.2 & 93.4$\pm$0.4 / 93.8$\pm$0.2 / \bf96.3$\pm$0.3\\
&& LSUN & 86.4$\pm$0.2 / 88.4$\pm$0.4 / \bf96.8$\pm$0.2 & 97.3$\pm$0.2 / 97.8$\pm$0.3 / \bf99.1$\pm$0.2 & 92.8$\pm$0.2 / 93.0$\pm$0.4 / \bf96.0$\pm$0.2\\
%%%%%%%%%%%%%%%%%%%%%%%%%%%%%%%%%%%%%%%%%%%%%%%%%%%%%%%%%%%%%%%%%%%%%%%%%%%%%%%%%%%%%%%
\midrule
\multirow{9}{*}{\begin{tabular}[c]{@{}c@{}}ResNet\end{tabular}}
& \multirow{3}{*}{\begin{tabular}[c]{@{}c@{}}CIFAR10\end{tabular}} 
& SVHN & 43.2$\pm$0.4 / 44.5$\pm$0.3 / \bf83.6$\pm$0.4 & 91.6$\pm$0.3 / 92.0$\pm$0.3 / \bf97.1$\pm$0.4 & 86.5$\pm$0.2 / 86.4$\pm$0.4 / \bf91.9$\pm$0.3\\
&& TinyImageNet & 46.4$\pm$0.4 / 48.0$\pm$0.3 / \bf70.3$\pm$0.3 & 89.8$\pm$0.4 / 90.1$\pm$0.2 / \bf94.6$\pm$0.4 & 84.0$\pm$0.4 / 84.2$\pm$0.3 / \bf88.3$\pm$0.4\\
&& LSUN & 51.2$\pm$0.4 / 53.3$\pm$0.2 / \bf82.3$\pm$0.3 & 92.2$\pm$0.4 / 92.6$\pm$0.4 / \bf96.9$\pm$0.3 & 86.5$\pm$0.2 / 86.6$\pm$0.4 / \bf91.5$\pm$0.3\\
\cmidrule{2-6} 
& \multirow{3}{*}{\begin{tabular}[c]{@{}c@{}}CIFAR100\end{tabular}} 
& SVHN & 15.9$\pm$0.8 / 18.0$\pm$0.7 / \bf20.2$\pm$0.6 & 71.3$\pm$0.6 / 72.7$\pm$0.7 / \bf85.3$\pm$0.6 & 66.1$\pm$0.6 / 66.3$\pm$0.7 / \bf79.7$\pm$0.6\\
&& TinyImageNet & 18.5$\pm$0.8 / 22.4$\pm$0.6 / \bf50.6$\pm$0.7 & 74.7$\pm$0.6 / 76.3$\pm$0.7 / \bf92.0$\pm$0.7 & 68.8$\pm$0.6 / 69.1$\pm$0.6 / \bf85.6$\pm$0.7\\
&& LSUN & 18.3$\pm$0.8 / 22.4$\pm$0.5 / \bf54.9$\pm$0.6 & 74.7$\pm$0.6 / 76.5$\pm$0.7 / \bf93.3$\pm$0.6 & 69.1$\pm$0.5 / 69.4$\pm$0.7 / \bf87.6$\pm$0.8\\
\cmidrule{2-6} 
& \multirow{3}{*}{\begin{tabular}[c]{@{}c@{}}SVHN\end{tabular}} 
& CIFAR10 & 67.3$\pm$0.2 / 67.7$\pm$0.3 / \bf92.3$\pm$0.2 & 89.8$\pm$0.2 / 89.7$\pm$0.3 / \bf98.0$\pm$0.2 & 87.0$\pm$0.3 / 86.9$\pm$0.3 / \bf94.1$\pm$0.2\\
&& TinyImageNet & 66.8$\pm$0.3 / 67.3$\pm$0.2 / \bf94.6$\pm$0.2 & 89.0$\pm$0.3 / 89.0$\pm$0.2 / \bf98.3$\pm$0.2 & 86.8$\pm$0.2 / 86.6$\pm$0.4 / \bf94.8$\pm$0.4\\
&& LSUN & 62.1$\pm$0.2 / 62.5$\pm$0.3 / \bf90.9$\pm$0.4 & 86.0$\pm$0.2 / 85.8$\pm$0.2 / \bf97.8$\pm$0.2 & 84.2$\pm$0.2 / 84.1$\pm$0.3 / \bf93.6$\pm$0.4\\
\bottomrule
\end{tabularx}
\vspace{-0.05in}
\begin{justify}The baseline OOD detection approaches are turnkey (no outlier/background/extra data required, no additional procedures other than typical straightforward neural network training are required), produce no classification accuracy drop, and present fast and energy-efficient inferences. Neither adversarial training, input preprocessing, temperature calibration, feature ensemble, nor metric learning is used. Since there is no need to tune hyperparameters, no access to OOD or adversarial examples is required. SoftMax+MPS means training with SoftMax loss and performing OOD detection using the maximum probability score, which is the approach defined in \cite{hendrycks2017baseline}. SoftMax+ES means training with SoftMax loss and performing OOD detection using the entropic score. IsoMax+ES means training with IsoMax loss and performing OOD detection using the entropic score. The OOD detection test sets are composed of images from both in-data and out-data (see Section~\ref{sec:experiments} for additional details). The results represent the mean and standard deviation of five executions. The best values are bold when they overcome the competing approach value outside of the margin of error given by the standard deviations. To the best of our knowledge, \emph{IsoMax+ES presents the state-or-the-art under these restrictive assumptions}.\end{justify}
\label{tbl:expanded_fair_odd}
\end{table*}

\begin{table*}%[t]
\centering
\caption{Comparison of Enhanced Versions of SoftMax Loss and IsoMax Loss:\\Adding Label Smoothing, Center Loss Regularization, ODIN, and Outlier Exposure to SoftMax Loss and IsoMax Loss.\\The Side Effects and Requirements Added to the Solution depend on the Add-on Technique.}\label{tbl:addon_ood}
\begin{tabularx}{\textwidth}{lll|CCC}
\toprule
%%%%%%%%%%%%%%%%%%%%%%%%%%%%%%%%%%%%%%%%%%%%%%%%%%%%%%%%%%%%%%%%%%%%%%%%%%%%%%%%%%%%%%%
%\multirow{4}{*}{\begin{tabular}[c]{@{}c@{}}\\Model\end{tabular}} & \multirow{4}{*}{\begin{tabular}[c]{@{}c@{}}\\In-Distribution\\(training)\end{tabular}} & \multirow{4}{*}{\begin{tabular}[c]{@{}c@{}}\\Out-Distribution\\(unseen)\end{tabular}} & 
\multirow{4}{*}{\begin{tabular}[c]{@{}c@{}}\\Model\end{tabular}} & \multirow{4}{*}{\begin{tabular}[c]{@{}c@{}}\\In-Data\\(training)\end{tabular}} & \multirow{4}{*}{\begin{tabular}[c]{@{}c@{}}\\OOD Detection\\Approach\end{tabular}} & 
%\multirow{4}{*}{\begin{tabular}[c]{@{}c@{}}\\OOD Detection Test Set\\(unseen)\end{tabular}} & 
%%%%%%%%%%%%%%%%%%%%%%%%%%%%%%%%%%%%%%%%%%%%%%%%%%%%%%%%%%%%%%%%%%%%%%%%%%%%%%%%%%%%%%%
%\multicolumn{3}{c}{Seamless Out-of-Distribution Detection:}\\
\multicolumn{3}{c}{Enhanced Versions of SoftMax loss and IsoMax loss.}\\
%&&& \multicolumn{3}{c}{Fast and Efficient Inferences. Scalable and Turnkey Approaches}\\
&&& \multicolumn{3}{c}{Add-on Techniques produce different Side Effects and Requirements.}\\
\cmidrule{4-6}
%&&& \multicolumn{3}{c}{Training Loss + Inference Score}\\
&&& Class. Accuracy (\%) [$\uparrow$] & TNR@TPR95 (\%) [$\uparrow$] & AUROC (\%) [$\uparrow$]\\
&&& SoftMax / IsoMax & SoftMax / IsoMax & SoftMax / IsoMax\\
%&&& TNR@TPR95 (\%) & AUROC (\%) & DTACC (\%)\\
%\cmidrule{4-6}
%&&& \multicolumn{3}{c}{Detection Metric}\\
%&&& \multicolumn{3}{c}{SoftMax+MPS \cite{hendrycks2017baseline} / SoftMax+ES / IsoMax+ES (ours)}\\
%%%%%%%%%%%%%%%%%%%%%%%%%%%%%%%%%%%%%%%%%%%%%%%%%%%%%%%%%%%%%%%%%%%%%%%%%%%%%%%%%%%%%%%
\midrule
\multirow{15}{*}{\begin{tabular}[c]{@{}c@{}}DenseNet\end{tabular}}
& \multirow{5}{*}{\begin{tabular}[c]{@{}c@{}}CIFAR10\end{tabular}} 
& Baseline & \bf{95.4$\pm$0.3} / \bf{95.2$\pm$0.3} & 53.3$\pm$0.4 / \bf{84.1$\pm$0.3} & 91.9$\pm$0.4 / \bf{97.3$\pm$0.3}\\
&& + Label Smoothing & \bf{95.2$\pm$0.4} / \bf{95.0$\pm$0.3} & \textbf{70.7$\pm$0.4} / 60.3$\pm$0.3 & \textbf{94.9$\pm$0.3} / 80.8$\pm$0.3\\
&& + Center Loss Regularization & \bf{95.3$\pm$0.3} / \bf{95.1$\pm$0.4} & 54.7$\pm$0.6 / \bf{87.1$\pm$0.3} & 92.8$\pm$0.3 / \bf{97.6$\pm$0.3}\\
&& + ODIN & \bf{95.4$\pm$0.3} / \bf{95.2$\pm$0.3} & 91.9$\pm$0.3 / {\color{blue}\bf{95.3$\pm$0.4}} & \bf{98.2$\pm$0.3} / \bf{98.4$\pm$0.4}\\
&& + Outlier Exposure & \bf{95.3$\pm$0.4} / \bf{95.6$\pm$0.4} & {\color{blue}93.8$\pm$0.3} / \bf{94.7$\pm$0.3} & {\color{blue}\bf{98.5$\pm$0.3}} / {\color{blue}\bf{98.8$\pm$0.4}}\\
\cmidrule{2-6} 
& \multirow{5}{*}{\begin{tabular}[c]{@{}c@{}}CIFAR100\end{tabular}} 
& Baseline & \bf{77.5$\pm$0.6} / \bf{77.5$\pm$0.4} & 22.3$\pm$0.7 / \bf{45.1$\pm$0.6} & 77.4$\pm$0.8 / \bf{91.9$\pm$0.6}\\
&& + Label Smoothing & \bf{77.0$\pm$0.4} / \bf{77.2$\pm$0.6} & 31.5$\pm$0.6 / \bf{33.0$\pm$0.6} & 82.0$\pm$0.6 / \bf{88.3$\pm$0.7}\\
&& + Center Loss Regularization & \bf{77.2$\pm$0.7} / \bf{77.0$\pm$0.6} & 30.3$\pm$0.7 / \bf{43.7$\pm$0.6} & 79.5$\pm$0.8 / \bf{92.2$\pm$0.6}\\
&& + ODIN & \bf{77.5$\pm$0.6} / \bf{77.5$\pm$0.4} & {\color{blue}64.4$\pm$0.7} / {\color{blue}\bf{83.1$\pm$0.8}} & {\color{blue}92.5$\pm$0.6} / {\color{blue}\bf{96.9$\pm$0.7}}\\
&& + Outlier Exposure & \bf{77.8$\pm$0.6} / \bf{77.5$\pm$0.7} & 23.0$\pm$0.7 / \bf{36.4$\pm$0.8} & 80.5$\pm$0.8 / \bf{89.6$\pm$0.6}\\
\cmidrule{2-6} 
& \multirow{5}{*}{\begin{tabular}[c]{@{}c@{}}SVHN\end{tabular}} 
& Baseline & \bf{96.6$\pm$0.2} / \bf{96.6$\pm$0.2} & 90.1$\pm$0.2 / \bf{95.9$\pm$0.1} & 98.2$\pm$0.2 / \bf{98.9$\pm$0.2}\\
&& + Label Smoothing & \bf{96.6$\pm$0.2} / \bf{96.7$\pm$0.3} & 87.5$\pm$0.2 / \bf{93.5$\pm$0.2} & 97.0$\pm$0.2 / \bf{97.8$\pm$0.3}\\
&& + Center Loss Regularization & \bf{96.7$\pm$0.3} / \bf{96.6$\pm$0.2} & 88.0$\pm$0.3 / \bf{95.9$\pm$0.2} & 97.9$\pm$0.2 / \bf{98.9$\pm$0.1}\\
&& + ODIN & \bf{96.6$\pm$0.2} / \bf{96.6$\pm$0.2} & 95.5$\pm$0.2 / \bf{96.7$\pm$0.2} & 98.8$\pm$0.1 / \bf{99.1$\pm$0.1}\\
&& + Outlier Exposure & \bf{96.6$\pm$0.3} / \bf{96.7$\pm$0.3} & {\color{blue}\bf{99.9$\pm$0.1}} / {\color{blue}\bf{99.9$\pm$0.1}} & {\color{blue}\bf{99.9$\pm$0.1}} / {\color{blue}\bf{99.9$\pm$0.1}}\\
\midrule
\multirow{15}{*}{\begin{tabular}[c]{@{}c@{}}ResNet\end{tabular}}
& \multirow{5}{*}{\begin{tabular}[c]{@{}c@{}}CIFAR10\end{tabular}} 
& Baseline & \bf{95.4$\pm$0.3} / \bf{95.6$\pm$0.4} & 50.8$\pm$0.4 / \bf{78.6$\pm$0.3} & 91.2$\pm$0.3 / \bf{96.1$\pm$0.4}\\
&& + Label Smoothing & \bf{95.5$\pm$0.4} / \bf{95.4$\pm$0.3} & 54.0$\pm$0.4 / \bf{63.5$\pm$0.3} & 78.9$\pm$0.4 / \bf{84.5$\pm$0.3}\\
&& + Center Loss Regularization & \bf{95.6$\pm$0.3} / \bf{95.4$\pm$0.4} & 53.5$\pm$0.4 / \bf{80.6$\pm$0.3} & 90.5$\pm$0.3 / \bf{96.6$\pm$0.2}\\
&& + ODIN & \bf{95.4$\pm$0.3} / \bf{95.6$\pm$0.4} & 73.7$\pm$0.3 / \bf{85.6$\pm$0.3} & 93.4$\pm$0.2 / \bf{97.2$\pm$0.3}\\
&& + Outlier Exposure & \bf{95.5$\pm$0.3} / \bf{95.6$\pm$0.3} & {\color{blue}91.1$\pm$0.3} / {\color{blue}\bf{94.2$\pm$0.4}} & {\color{blue}97.7$\pm$0.2} / {\color{blue}\bf{98.6$\pm$0.3}}\\
\cmidrule{2-6} 
& \multirow{5}{*}{\begin{tabular}[c]{@{}c@{}}CIFAR100\end{tabular}} 
& Baseline & \bf{77.3$\pm$0.6} / \bf{77.4$\pm$0.7} & 22.4$\pm$0.6 / \bf{41.8$\pm$0.7} & 80.5$\pm$0.6 / \bf{90.1$\pm$0.7}\\
&& + Label Smoothing & \bf{77.7$\pm$0.5} / \bf{77.3$\pm$0.5} & 21.0$\pm$0.7 / \bf{33.4$\pm$0.6} & 81.6$\pm$0.6 / \bf{85.9$\pm$0.6}\\
&& + Center Loss Regularization & \bf{77.9$\pm$0.5} / \bf{77.4$\pm$0.5} & 29.5$\pm$0.9 / \bf{48.2$\pm$0.7} & 80.3$\pm$0.6 / \bf{90.6$\pm$0.6}\\
&& + ODIN & \bf{77.3$\pm$0.6} / \bf{77.4$\pm$0.7} & {\color{blue}64.0$\pm$0.6} / {\color{blue}\bf{77.9$\pm$0.6}} & {\color{blue}92.7$\pm$0.7} / {\color{blue}\bf{95.8$\pm$0.6}}\\
&& + Outlier Exposure & \bf{77.3$\pm$0.5} / \bf{77.0$\pm$0.5} & 37.8$\pm$0.9 / \bf{41.7$\pm$0.8} & 86.6$\pm$0.6 / \bf{88.6$\pm$0.7}\\
\cmidrule{2-6} 
& \multirow{5}{*}{\begin{tabular}[c]{@{}c@{}}SVHN\end{tabular}} 
& Baseline & \bf{96.8$\pm$0.2} / \bf{96.7$\pm$0.2} & 71.7$\pm$0.3 / \bf{92.4$\pm$0.2} & 94.7$\pm$0.3 / \bf{98.0$\pm$0.2}\\
&& + Label Smoothing & \bf{96.9$\pm$0.2} / \bf{96.9$\pm$0.3} & \bf{86.3$\pm$0.2} / \bf{86.4$\pm$0.2} & \textbf{97.2$\pm$0.3} / 94.9$\pm$0.2\\
&& + Center Loss Regularization & \bf{96.9$\pm$0.2} / \bf{96.7$\pm$0.2} & 77.2$\pm$0.2 / \bf{82.2$\pm$0.2} & 94.5$\pm$0.2 / \bf{96.1$\pm$0.3}\\
&& + ODIN & \bf{96.8$\pm$0.2} / \bf{96.7$\pm$0.2} & 79.9$\pm$0.3 / \bf{93.5$\pm$0.3} & 95.1$\pm$0.2 / \bf{98.2$\pm$0.3}\\
&& + Outlier Exposure & \bf{96.9$\pm$0.3} / \bf{96.8$\pm$0.2} & {\color{blue}\bf{99.9$\pm$0.1}} / {\color{blue}\bf{99.9$\pm$0.1}} & {\color{blue}\bf{99.9$\pm$0.1}} / {\color{blue}\bf{99.9$\pm$0.1}}\\
\bottomrule
\end{tabularx}
\begin{justify}
Regardless of using SoftMax loss or IsoMax loss, adding label smoothing (LS) requires validation of the hyperparameter $\epsilon$ \cite{DBLP:conf/cvpr/SzegedyVISW16}. The values searched for $\epsilon$ were 0.1 \cite{DBLP:conf/cvpr/SzegedyVISW16, DBLP:journals/corr/abs-2007-03212} and 0.01 \cite{DBLP:journals/corr/abs-2007-03212}. Adding center loss regularization (CLR) to SoftMax loss or IsoMax loss requires validation of the hyperparameter $\lambda$ \cite{DBLP:conf/eccv/WenZL016}. The values searched for $\lambda$ were 0.01 and 0.003 \cite{DBLP:conf/eccv/WenZL016}. We emphasize that the center loss is not used as a stand-alone loss but rather combined as a regularization term with a preexisting baseline loss. In the case of the original paper, the baseline loss was the SoftMax loss. In this paper, we added the mentioned regularization term \cite[Equation (5)]{DBLP:conf/eccv/WenZL016} also to IsoMax loss to construct the center loss enhanced version of our loss. Regardless of using SoftMax loss or IsoMax loss, adding ODIN \cite{liang2018enhancing} requires validation of the hyperparameters $\epsilon$ and $T$. The values searched for these hyperparameters were the same used in the original paper \cite{liang2018enhancing}. Adding ODIN implies using input preprocessing, which makes inferences much slower, and energy- and cost-inefficient. We used OOD data to validate the LS, CLR, and ODIN hyperparameters. Adding outlier exposure (OE) \cite{hendrycks2018deep} to SoftMax loss or IsoMax loss requires collecting outlier data. We used the same outlier data used in \cite{hendrycks2018deep}. The add-on techniques were not applied to the SoftMax and IsoMax losses combined, but rather individually. The values of the performance metrics TNR@TPR95 and AUROC were averaged over all out-of-distribution data. All results used the entropic score, as it always overcame the maximum probability score. The results represent the mean and standard deviation of five executions. The best values are bold when they overcome the competing approach value outside of the margin of error given by the standard deviations. {\color{black}For OOD detection, the variants of SoftMax and IsoMax losses that presented the best performance for each combination of architecture and dataset are blue.} The baseline results are in blue.
\end{justify}
\label{tbl:addon_odd}
\end{table*}

\begin{table*}%[t]
%\color{blue}
%\renewcommand{\arraystretch}{0.5}
\caption{Baseline Out-of-Distribution Detection Approaches Comparison: Text Data}\label{tbl:text_odd}
%\small
\centering
%\vspace{-0.12in}
%\begin{tabularx}{\textwidth}{CCC|CC}
\begin{tabularx}{\textwidth}{lll|CC}
\toprule
%%%%%%%%%%%%%%%%%%%%%%%%%%%%%%%%%%%%%%%%%%%%%%%%%%%%%%%%%%%%%%%%%%%%%%%%%%%%%%%%%%%%%%%
%\multirow{4}{*}{\begin{tabular}[c]{@{}c@{}}\\Model\end{tabular}} & \multirow{4}{*}{\begin{tabular}[c]{@{}c@{}}\\In-Distribution\\(training)\end{tabular}} & \multirow{4}{*}{\begin{tabular}[c]{@{}c@{}}\\Out-Distribution\\(unseen)\end{tabular}} & 
\multirow{4}{*}{\begin{tabular}[c]{@{}c@{}}\\Model\end{tabular}} & \multirow{4}{*}{\begin{tabular}[c]{@{}c@{}}\\In-Data\\(training)\end{tabular}} & %\multirow{4}{*}{\begin{tabular}[c]{@{}c@{}}\\Out-Data\\(unseen)\end{tabular}} & 
\multirow{4}{*}{\begin{tabular}[c]{@{}c@{}}\\Out-Data\\(unseen)\end{tabular}} & 
%%%%%%%%%%%%%%%%%%%%%%%%%%%%%%%%%%%%%%%%%%%%%%%%%%%%%%%%%%%%%%%%%%%%%%%%%%%%%%%%%%%%%%%
\multicolumn{2}{c}{Baseline Out-of-Distribution Detection Approaches Comparison.}\\
&&& \multicolumn{2}{c}{Fast and Energy-Efficient Inferences. Turnkey Approaches. No outlier data used.}\\
%&&& \multicolumn{2}{c}{Out-of-Distribution Detection}\\
%&&& \multicolumn{2}{c}{Training Loss + Inference Score}\\
\cmidrule{4-5}
&&& FPR@TPR90 (\%) [$\downarrow$] & AUROC (\%) [$\uparrow$]\\
%\cmidrule{4-5}
%&&& \multicolumn{2}{c}{Detection Metric}\\
&&& \multicolumn{2}{c}{SoftMax+MPS \cite{hendrycks2017baseline} / IsoMax+ES (ours)}\\
%%%%%%%%%%%%%%%%%%%%%%%%%%%%%%%%%%%%%%%%%%%%%%%%%%%%%%%%%%%%%%%%%%%%%%%%%%%%%%%%%%%%%%%
\midrule
\multirow{3}{*}{\begin{tabular}[c]{@{}c@{}}GRU2L~~~~\end{tabular}}
& \multirow{3}{*}{\begin{tabular}[c]{@{}c@{}}20 Newsgroups~~~~\end{tabular}} 
%& SNLI & 72.7$\pm$0.3 / \bf74.5$\pm$0.3 & {\bf33.1$\pm$0.4} / \bf32.7$\pm$0.5\\
& IMDB~~~~& 34.9$\pm$0.3 / \bf22.7$\pm$0.4 & 85.4$\pm$0.3 / \bf90.0$\pm$0.2\\
&& Multi30K & 47.2$\pm$0.4 / \bf44.5$\pm$0.3 & 80.9$\pm$0.4 / \bf82.7$\pm$0.4\\
&& Yelp Reviews~~~~~~~~& 39.3$\pm$0.3 / \bf26.1$\pm$0.6 & 82.5$\pm$0.4 / \bf88.2$\pm$0.4\\
%\cmidrule{2-5} 
\bottomrule
\end{tabularx}
\begin{justify}%The baseline OOD detection approaches are turnkey (no outlier/background/extra data required, no additional procedures other than typical straightforward neural network training are required), produce no classification accuracy drop, and present fast and energy-efficient inferences. Neither adversarial training, input preprocessing, temperature calibration, feature ensemble, nor metric learning is used. Since there is no need to tune hyperparameters, no access to OOD or adversarial examples is required. 
%SoftMax+MPS means training with SoftMax loss and performing OOD detection using the maximum probability score, which is the approach defined in \cite{hendrycks2017baseline}. 
\textcolor{black}{SoftMax+MPS means training with SoftMax loss and performing OOD detection using the maximum probability score. IsoMax+ES means training with IsoMax loss and performing OOD detection using the entropic score. The results represent the mean and standard deviation of five executions. The best values are bold when they overcome the competing approach value outside of the margin of error given by the standard deviations.}
%To the best of our knowledge, \emph{IsoMax+ES presents the state-or-the-art under these restrictive assumptions}.
\end{justify}
%\vskip 0.25cm
\end{table*}

\begin{table*}%[t]
%\color{blue}
\renewcommand{\arraystretch}{0.5}
\caption{Comparison of the Proposed Baseline Out-of-Distribution Detection Approach with No Seamless Solutions.\\Unfair Comparison of Approaches with Different Requirements and Side Effects. No extra/outlier/background data used.}\label{tbl:unfair_odd}
%\small
\centering
%\vspace{-0.12in}
%\begin{tabularx}{\textwidth}{CCC|CC}
\begin{tabularx}{\textwidth}{lll|CC}
\toprule
%%%%%%%%%%%%%%%%%%%%%%%%%%%%%%%%%%%%%%%%%%%%%%%%%%%%%%%%%%%%%%%%%%%%%%%%%%%%%%%%%%%%%%%
%\multirow{4}{*}{\begin{tabular}[c]{@{}c@{}}\\Model\end{tabular}} & \multirow{4}{*}{\begin{tabular}[c]{@{}c@{}}\\In-Distribution\\(training)\end{tabular}} & \multirow{4}{*}{\begin{tabular}[c]{@{}c@{}}\\Out-Distribution\\(unseen)\end{tabular}} & 
\multirow{4}{*}{\begin{tabular}[c]{@{}c@{}}\\Model\end{tabular}} & \multirow{4}{*}{\begin{tabular}[c]{@{}c@{}}\\In-Data\\(training)\end{tabular}} & %\multirow{4}{*}{\begin{tabular}[c]{@{}c@{}}\\Out-Data\\(unseen)\end{tabular}} & 
\multirow{4}{*}{\begin{tabular}[c]{@{}c@{}}\\Out-Data\\(unseen)\end{tabular}} & 
%%%%%%%%%%%%%%%%%%%%%%%%%%%%%%%%%%%%%%%%%%%%%%%%%%%%%%%%%%%%%%%%%%%%%%%%%%%%%%%%%%%%%%%
\multicolumn{2}{c}{ODIN, ACET, and Mahalanobis present troublesome requirements.}\\
&&& \multicolumn{2}{c}{ODIN, ACET, and Mahalanobis produce undesired side effects.}\\
%&&& \multicolumn{2}{c}{Out-of-Distribution Detection}\\
%&&& \multicolumn{2}{c}{Training Loss + Inference Score}\\
\cmidrule{4-5}
&&& AUROC (\%) [$\uparrow$] & DTACC (\%) [$\uparrow$]\\
%\cmidrule{4-5}
%&&& \multicolumn{2}{c}{Detection Metric}\\
&&& \multicolumn{2}{c}{ODIN \cite{liang2018enhancing} / ACET \cite{Hein2018WhyRN} / IsoMax+ES (ours) / Mahalanobis \cite{lee2018simple}}\\
%%%%%%%%%%%%%%%%%%%%%%%%%%%%%%%%%%%%%%%%%%%%%%%%%%%%%%%%%%%%%%%%%%%%%%%%%%%%%%%%%%%%%%%
\midrule
\multirow{9}{*}{\begin{tabular}[c]{@{}c@{}}DenseNet~~~~\end{tabular}}
& \multirow{3}{*}{\begin{tabular}[c]{@{}c@{}}CIFAR10~~~~\end{tabular}} 
& SVHN & 92.7$\pm$0.4 / NA / {\bf96.7$\pm$0.4} / \bf97.5$\pm$0.6 & 86.4$\pm$0.4 / NA / {\bf91.8$\pm$0.4} / \bf92.4$\pm$0.5\\
&& TinyImageNet~~~~& 97.3$\pm$0.4 / NA / {\bf97.9$\pm$0.4} / \bf98.5$\pm$0.6 & 92.1$\pm$0.4 / NA / {\bf93.5$\pm$0.4} / \bf94.3$\pm$0.6\\
&& LSUN & 98.4$\pm$0.4 / NA / {\bf98.9$\pm$0.4} / \bf99.1$\pm$0.6 & 94.3$\pm$0.3 / NA / {\bf95.1$\pm$0.4} / \bf95.9$\pm$0.4\\
\cmidrule{2-5} 
& \multirow{3}{*}{\begin{tabular}[c]{@{}c@{}}CIFAR100\end{tabular}} 
& SVHN & 88.1$\pm$0.6 / NA / 88.7$\pm$0.7 / \bf91.7$\pm$0.6 & 80.8$\pm$0.6 / NA / {\bf83.9$\pm$0.6} / \bf84.3$\pm$0.7\\
&& TinyImageNet & 85.2$\pm$0.6 / NA / 92.8$\pm$0.5 / \bf96.9$\pm$0.7 & 77.1$\pm$0.6 / NA / 86.7$\pm$0.5 / \bf91.7$\pm$0.6\\
&& LSUN & 85.8$\pm$0.6 / NA / 94.6$\pm$0.4 / \bf97.7$\pm$0.6 & 77.3$\pm$0.6 / NA / 89.2$\pm$0.7 / \bf93.5$\pm$0.5\\
\cmidrule{2-5} 
& \multirow{3}{*}{\begin{tabular}[c]{@{}c@{}}SVHN\end{tabular}} 
& CIFAR10 & 91.8$\pm$0.2 / NA / {\bf98.6$\pm$0.2} / \bf98.7$\pm$0.3 & 86.7$\pm$0.2 / NA / {\bf95.7$\pm$0.3} / \bf96.1$\pm$0.3\\
&& TinyImageNet & 94.8$\pm$0.2 / NA / {\bf99.3$\pm$0.2} / \bf99.7$\pm$0.3 & 90.2$\pm$0.2 / NA / 96.2$\pm$0.2 / \bf98.8$\pm$0.3\\
&& LSUN & 94.0$\pm$0.2 / NA / {\bf99.5$\pm$0.2} / \bf99.8$\pm$0.3 & 89.1$\pm$0.2 / NA / 96.1$\pm$0.2 / \bf99.0$\pm$0.1\\
%%%%%%%%%%%%%%%%%%%%%%%%%%%%%%%%%%%%%%%%%%%%%%%%%%%%%%%%%%%%%%%%%%%%%%%%%%%%%%%%%%%%%%%
\midrule
\multirow{9}{*}{\begin{tabular}[c]{@{}c@{}}ResNet\end{tabular}}
& \multirow{3}{*}{\begin{tabular}[c]{@{}c@{}}CIFAR10\end{tabular}} 
& SVHN & 86.4$\pm$0.4 / {\bf97.7$\pm$0.4} / {\bf97.4$\pm$0.4} / 95.5$\pm$0.6 & 77.7$\pm$0.4 / NA / {\bf91.8$\pm$0.4} / 89.0$\pm$0.5\\
&& TinyImageNet & 93.9$\pm$0.3 / 85.7$\pm$0.4 / 94.7$\pm$0.3 / \bf99.1$\pm$0.5 & 86.1$\pm$0.3 / NA / 88.5$\pm$0.3 / \bf95.4$\pm$0.5\\
&& LSUN & 93.4$\pm$0.4 / 85.9$\pm$0.4 / 96.8$\pm$0.3 / \bf99.5$\pm$0.3 & 85.7$\pm$0.4 / NA / 91.3$\pm$0.3 / \bf97.3$\pm$0.4\\
\cmidrule{2-5} 
& \multirow{3}{*}{\begin{tabular}[c]{@{}c@{}}CIFAR100\end{tabular}} 
& SVHN & 72.1$\pm$0.6 / {\bf91.1$\pm$0.6} / 85.2$\pm$0.6 / 84.3$\pm$0.5 & 67.8$\pm$0.4 / NA / {\bf79.9$\pm$0.6} / 76.4$\pm$0.7\\
&& TinyImageNet & 83.7$\pm$0.6 / 75.3$\pm$0.5 / {\bf92.4$\pm$0.5} / 87.7$\pm$0.6 & 75.7$\pm$0.5 / NA / {\bf85.8$\pm$0.5} / 84.3$\pm$0.5\\
&& LSUN & 81.8$\pm$0.6 / 69.7$\pm$0.5 / {\bf93.3$\pm$0.6} / 82.2$\pm$0.7 & 74.7$\pm$0.6 / NA / {\bf87.6$\pm$0.5} / 79.6$\pm$0.6\\
\cmidrule{2-5} 
& \multirow{3}{*}{\begin{tabular}[c]{@{}c@{}}SVHN\end{tabular}} 
& CIFAR10 & 92.1$\pm$0.2 / 97.3$\pm$0.3 / {\bf98.2$\pm$0.2} / 97.3$\pm$0.2 & 89.3$\pm$0.3 / NA / {\bf94.3$\pm$0.2} / \bf94.5$\pm$0.3\\
&& TinyImageNet & 92.8$\pm$0.2 / 97.6$\pm$0.2 / {\bf98.8$\pm$0.1} / \bf99.0$\pm$0.3 & 90.0$\pm$0.2 / NA / 94.6$\pm$0.3 / \bf98.7$\pm$0.2\\
&& LSUN & 90.6$\pm$0.2 / {\bf99.7$\pm$0.3} / 97.9$\pm$0.2 / \bf99.8$\pm$0.2 & 88.3$\pm$0.2 / NA / 93.7$\pm$0.3 / \bf99.4$\pm$0.2\\
\bottomrule
\end{tabularx}
\begin{justify}ODIN uses input preprocessing, temperature calibration, and adversarial validation (hyperparameter tuning using adversarial examples). The Mahalanobis solution uses input preprocessing, adversarial validation, feature extraction, feature ensemble, and metric learning. Input preprocessing makes the inferences of ODIN and the Mahalanobis method at least three times slower and \emph{at least three times less energy/computationally efficient than SoftMax or IsoMax inferences}. Feature ensembles may limit the Mahalanobis method scalability to deal with large-size images used in real-world applications. ACET uses adversarial training, which results in slower training, possibly reduced scalability for large-size images, and eventually, classification accuracy drop. The ODIN, the Mahalanobis approach, and ACET present hyperparameters that need to be validated for each combination of datasets and models presented in the table. Furthermore, considering that adversarial hyperparameters (e.g., the adversarial perturbation) used in ODIN/Mahalanobis/ACET were validated using the SVHN/CIFAR10/CIFAR100 validation sets and that these sets were reused as OOD detection test set, we conclude that the {\color{black}OOD} detection performance reported by those papers, which we are reproducing in this table, may be overestimated. ODIN/Mahalanobis/ACET results were obtained using SoftMax loss rather than IsoMax loss as baseline. %Hyperparameter validation also requires in-distribution validation samples, which should be removed from training data for a fair comparison. In \cite{lee2018simple}, these samples were removed from the validation set with the remaining images used for the test. Therefore, ODIN and the Mahalanobis method results were obtained in a more favorable training condition than IsoMax, which did not use these extra samples for training.
No outlier/extra/background data is used. IsoMax+ES means training with IsoMax loss and performing OOD detection using the entropic score. IsoMax+ES does not use these techniques and does not have such requirements, side effects, and hyperparameters to tune (as IsoMax+ES works as a baseline OOD detection approach, those techniques may be incorporated in future research). The OOD detection test sets are composed of images from both in-data and out-data (see Section~\ref{sec:experiments} for further details). %Editor: Please ensure that the intended meaning has been maintained in the edits in the previous sentence. 
The results represent the mean and standard deviation of five executions. The best values are bold when they overcome the competing approach value outside the margin of error given by the standard deviations.\end{justify}
%\vskip 0.1cm
\caption{Out-of-Distribution Detection: Inference Delays / Presumed Computational Cost and Energy Consumption Rates.}\label{tbl:times}
\begin{tabularx}{\textwidth}{lll|CCC}
\toprule
%%%%%%%%%%%%%%%%%%%%%%%%%%%%%%%%%%%%%%%%%%%%%%%%%%%%%%%%%%%%%%%%%%%%%%%%%%%%%%%%%%%%%%%
%\multirow{4}{*}{\begin{tabular}[c]{@{}c@{}}\\Model\end{tabular}} & \multirow{4}{*}{\begin{tabular}[c]{@{}c@{}}\\In-Distribution\\(training)\end{tabular}} & \multirow{4}{*}{\begin{tabular}[c]{@{}c@{}}\\Out-Distribution\\(unseen)\end{tabular}} & 
\multirow{4}{*}{\begin{tabular}[c]{@{}c@{}}\\Model\end{tabular}} & \multirow{4}{*}{\begin{tabular}[c]{@{}c@{}}\\In-Data\\(training)\end{tabular}} & \multirow{4}{*}{\begin{tabular}[c]{@{}c@{}}\\Hardware\\(inference)\end{tabular}} & 
%\multirow{4}{*}{\begin{tabular}[c]{@{}c@{}}\\OOD Detection Test Set\\(unseen)\end{tabular}} & 
%%%%%%%%%%%%%%%%%%%%%%%%%%%%%%%%%%%%%%%%%%%%%%%%%%%%%%%%%%%%%%%%%%%%%%%%%%%%%%%%%%%%%%%
%\multicolumn{3}{c}{Seamless Out-of-Distribution Detection:}\\
\multicolumn{3}{c}{Out-of-Distribution Detection:}\\
%&&& \multicolumn{3}{c}{Fast and Efficient Inferences. Scalable and Turnkey Approaches}\\
&&& \multicolumn{3}{c}{Inference Delays / Presumed Computational Cost and Energy Consumption Rates.}\\
\cmidrule{4-6}
%&&& \multicolumn{3}{c}{Training Loss + Inference Score}\\
&&& SoftMax Loss \cite{hendrycks2017baseline} \mbox{(current baseline)} & IsoMax Loss (ours) \mbox{(proposed baseline)} & Input Preprocessing: \mbox{ODIN \cite{liang2018enhancing}, Mahalanobis \cite{lee2018simple}}, Generalized ODIN \cite{Hsu2020GeneralizedOD}\\
&&& MPS (ms) [$\downarrow$] / ES (ms) [$\downarrow$] & MPS (ms) [$\downarrow$] / ES (ms) [$\downarrow$] & (ms) [$\downarrow$]\\
%&&& TNR@TPR95 (\%) & AUROC (\%) & DTACC (\%)\\
%\cmidrule{4-6}
%&&& \multicolumn{3}{c}{Detection Metric}\\
%&&& \multicolumn{3}{c}{SoftMax+MPS \cite{hendrycks2017baseline} / SoftMax+ES / IsoMax+ES (ours)}\\
%%%%%%%%%%%%%%%%%%%%%%%%%%%%%%%%%%%%%%%%%%%%%%%%%%%%%%%%%%%%%%%%%%%%%%%%%%%%%%%%%%%%%%%
\midrule
\multirow{6}{*}{\begin{tabular}[c]{@{}c@{}}DenseNet~~~~~~\end{tabular}}
& \multirow{2}{*}{\begin{tabular}[c]{@{}c@{}}CIFAR10~~~~~~~\end{tabular}} 
&~~~~CPU~~~~~~& 18.1 / 19.4 & 18.0 / 19.2 & 242.4 \bf{($\approx$ 10x slower)}\\
&&~~~~GPU~~~~~~& 11.6 / 13.0 & 11.6 / 11.5 & 39.2 \bf{($\approx$ 4x slower)}\\
\cmidrule{2-6} 
& \multirow{2}{*}{\begin{tabular}[c]{@{}c@{}}CIFAR100\end{tabular}} 
&~~~~CPU~~~~~~& 18.4 / 19.8 & 18.4 / 19.3 & 261.0 \bf{($\approx$ 10x slower)}\\
&&~~~~GPU~~~~~~& 12.9 / 11.4 & 11.8 / 11.5 & 39.6 \bf{($\approx$ 4x slower)}\\
\cmidrule{2-6} 
& \multirow{2}{*}{\begin{tabular}[c]{@{}c@{}}SVHN\end{tabular}} 
&~~~~CPU~~~~~~& 18.1 / 18.6 & 18.3 / 18.6 & 241.5 \bf{($\approx$ 10x slower)}\\
&&~~~~GPU~~~~~~& 11.6 / 11.9 & 11.7 / 11.6 & 39.6 \bf{($\approx$ 4x slower)}\\
%%%%%%%%%%%%%%%%%%%%%%%%%%%%%%%%%%%%%%%%%%%%%%%%%%%%%%%%%%%%%%%%%%%%%%%%%%%%%%%%%%%%%%%
\midrule
\multirow{6}{*}{\begin{tabular}[c]{@{}c@{}}ResNet\end{tabular}}
& \multirow{2}{*}{\begin{tabular}[c]{@{}c@{}}CIFAR10\end{tabular}} 
&~~~~CPU~~~~~~& 22.3 / 23.2 & 23.0 / 23.5 & 250.4 \bf{($\approx$ 10x slower)}\\
&&~~~~GPU~~~~~~& 4.5 / 3.8 & 4.2 / 4.1 & 15.4 \bf{($\approx$ 4x slower)}\\
\cmidrule{2-6} 
& \multirow{2}{*}{\begin{tabular}[c]{@{}c@{}}CIFAR100\end{tabular}} 
&~~~~CPU~~~~~~& 23.3 / 23.1 & 23.3 / 23.8 & 252.6 \bf{($\approx$ 10x slower)}\\
&&~~~~GPU~~~~~~& 4.3 / 3.9 & 4.3 / 4.2 & 14.8 \bf{($\approx$ 4x slower)}\\
\cmidrule{2-6} 
& \multirow{2}{*}{\begin{tabular}[c]{@{}c@{}}SVHN\end{tabular}} 
&~~~~CPU~~~~~~& 23.1 / 23.4 & 23.4 / 23.3 & 263.8 \bf{($\approx$ 10x slower)}\\
&&~~~~GPU~~~~~~& 4.2 / 4.0 & 4.0 / 4.0 & 15.7 \bf{($\approx$ 4x slower)}\\
\bottomrule
\end{tabularx}
\begin{justify}MPS means maximum probability score. ES means entropic score. For SoftMax and IsoMax losses (baseline OOD detection approaches), the inference delays combine both classification and detection computation. For the methods based on input preprocessing, the inference delays represent only the input preprocessing phase. All values are in milliseconds. In addition to presenting similar classification accuracy (Table~\ref{tbl:addon_ood}, Supplementary Material~\ref{apx:robustness_study}) and much better OOD detection performance (Table~\ref{tbl:expanded_fair_odd}), IsoMax loss trained networks produce inferences as fast as SoftMax trained networks. Moreover, the entropic score is as fast as the maximum probability score. Using CPU (Intel i7-4790K, 4.00GHz, x64, octa-core) for inference (the case more relevant from a cost point of view), methods based on input preprocessing are \emph{about ten times slower} than the baseline approaches. Using GPU (Nvidia GTX 1080 Ti) for inference, our approach is \emph{about four times faster} than the methods based on input preprocessing. The inference delay rates presumably reflect in similar the computational cost and energy consumption rates. The inference delays presented are the mean value of the inference delay of each image calculated (batch size equals one) over the entire dataset. The standard deviation was below 0.3 for all cases.\end{justify}
%\label{tbl:unfair_odd}
\end{table*}

All datasets, models, training procedures, and evaluation metrics followed the baseline established in \cite{hendrycks2017baseline} and were subsequently used in major OOD detection papers \cite{liang2018enhancing,lee2018simple,Hein2018WhyRN,hendrycks2018deep}. We only compared approaches that did not present classification accuracy drop in the experiments presented in the work in which they were proposed. The \emph{code} is available\footnote{\url{https://github.com/dlmacedo/entropic-out-of-distribution-detection}}.

We trained from scratch 100-layer DenseNet-BC \cite{Huang2017DenselyNetworks} (growth rate $k\!\!=\!12$, 0.8M parameters) and 34-layer ResNets \cite{He_2016} on CIFAR10 \cite{Krizhevsky2009LearningImages}, CIFAR100 \cite{Krizhevsky2009LearningImages} and SVHN \cite{Netzer2011ReadingLearning} using SoftMax and IsoMax losses. We used SGD with the Nesterov moment equal to 0.9, 300 epochs with a batch size of 64, and an initial learning rate of 0.1 with a learning rate decay rate equal to ten applied in epochs 150, 200, and 250. We used dropout equal to zero. The weight decay was 0.0001. We used resized images from the TinyImageNet \cite{Deng2009ImageNetDatabase}\footnote{\label{odin_code}\url{https://github.com/facebookresearch/odin}} and the Large-scale Scene UNderstanding (LSUN) \cite{Yu2015LSUNLoop}\textsuperscript{\ref{odin_code}} datasets as OOD data. 
%Resized images from \cite{NguyenYC15}\footnote{\url{http://anhnguyen.me/project/fooling}} compose the fooling OOD data. The Gaussian OOD data is composed of images in which every pixel is independently and identically sampled from a Gaussian distribution with a mean of 0.5 and unit variance with values clipped into the range [0, 1]. The uniform OOD data is formed by images in which every pixel is independently and identically sampled from the uniform distribution on~[0,~1]. 
Each set of OOD data presents ten thousand images.

Finally, we separately added these OOD data to the CIFAR10, CIFAR100, and SVHN validation sets to construct the respective OOD detection test sets. For example, the OOD detection test set for in-distribution CIFAR10 and out-distribution TinyImageNet is composed by combing the CIFAR validation set and the TinyImageNet OOD data. In some cases, CIFAR10 and SVHN work as out-distribution and the respective validation set is used to construct the OOD detection test set. %For example, the OOD detection test set for in-distribution CIFAR100 and out-distribution SVHN is composed by combing their dataset validation sets. 
We emphasize that our solution does not use validation sets for training or validation. Therefore, validation sets are used exclusively for building OOD detection test sets.

\textcolor{black}{For text data experiments, we followed the experimental setting presented in \cite{hendrycks2018deep}. Therefore, we used the 20 Newsgroups as in-distribution data. The 20 Newsgroups is a text classification dataset of newsgroup documents. It has 20 classes and approximately 20,000 examples that are split evenly among the classes. We used the standard 60/40 train/test split. We used the IMDB, Multi30K, and Yelp Reviews datasets as out-distribution. IMDB is a dataset of movie review sentiment classification. Multi30K is a dataset of English-German image descriptions, of which we use the English descriptions. Yelp Reviews is a dataset of restaurant reviews. Finally, we trained from scratch 2-layer GRUs (GRU2L) \cite{DBLP:conf/emnlp/ChoMGBBSB14} using SoftMax and IsoMax losses.}

%{\color{blue}, the area under the precision-recall curve (AUPR),} 
We evaluated the performance using three detection metrics. First, we calculated the true negative rate at the 95\% true positive rate (TNR@TPR95) {\color{black}and the false positive rate at the 90\% true positive rate (FPR@TPR90)} using the adequate thresholds. In addition, we evaluated the area under the receiver operating characteristic curve (AUROC)and the detection accuracy (DTACC), which corresponds to the maximum classification probability over all possible thresholds $\delta$:
\begin{align}
\begin{split}
1- \min_{\delta} \big\{ P_{\texttt{in}} \left( o \left( \mathbf{x} \right) \leq \delta \right) P \left(\mathbf{x}\text{ is from }P_{\texttt{in}}\right)
\\+ P_{\texttt{out}} \left( o \left( \mathbf{x} \right) >\delta \right) P \left(\mathbf{x}\text{ is from }P_{\texttt{out}}\right)\big\},
\end{split}
\end{align}
where $o(\mathbf{x})$ is a given OOD detection score. It is assumed that both positive and negative samples have equal probability. % of being in the test set. %, i.e., $P \left(\mathbf{x}\text{ is from }P_{\texttt{in}}\right) = P\left(\mathbf{x}\text{ is from }P_{\texttt{out}}\right)$.
%All metrics follow the calculation detailed in \cite{lee2018simple}.
In other words, the DTACC represents the probability that we can correctly classify whether a sample belongs to the in-distribution or the out-distribution considering the best case scenario, i.e., using the ideal value for $\delta$. AUROC and DTACC are threshold independent. %{\color{blue}, AUPR,}

\subsection{Entropic Scale, High Entropy, and OOD Detection}

To experimentally show that higher entropic scales lead to higher entropy probability distributions and consequently improved OOD detection performance, we trained DenseNets on SVHN using the SoftMax loss and IsoMax loss with distinct entropic scale values. We used the entropic score and the TNR@TPR95 to evaluate OOD detection performance (please, see Fig.~\ref{fig:train_losses_entropies_and_entropic_scale_parametrization}).

As expected, Fig.~\ref{fig:train_losses_entropies_and_entropic_scale_parametrization}a shows that the SoftMax loss generates posterior distributions with low entropy. Fig.~\ref{fig:train_losses_entropies_and_entropic_scale_parametrization}b illustrates that the unitary entropic scale \mbox{($E_s\!\!=\!1$)} only slightly increases the distribution mean entropy. In other words, isotropy alone is not enough to circumvent the cross-entropy propensity to produce low entropy probability distributions and the entropic scale is indeed necessary. Nevertheless, the replacement of anisotropic logits based on affine transformation by isotropic logits is enough to produce initial OOD detection performance gains regardless of the out-distribution (Fig.~\ref{fig:train_losses_entropies_and_entropic_scale_parametrization}e).

Training using \mbox{$E_s\!\!=\!1$} and then making during inference the temperature \mbox{$T\!\!=\!0.1$}, which is equivalent to make \mbox{$E_s\!\!=\!10$}, produces different results from training using \mbox{$E_s\!\!=\!10$} and removing it for inference. Indeed, in the first case we only change the last layer, while in the second case, we learn different weights for the entire neural network in comparison with training with \mbox{$E_s\!\!=\!1$} (Fig.~\ref{fig:train_losses_entropies_and_entropic_scale_parametrization}e).

Fig.~\ref{fig:train_losses_entropies_and_entropic_scale_parametrization}c shows that an intermediate entropic scale \mbox{($E_s\!\!=\!3$)} provides medium entropy probability distributions with corresponding additional OOD detection performance gains for all out-distributions (Fig.~\ref{fig:train_losses_entropies_and_entropic_scale_parametrization}e). Fig.~\ref{fig:train_losses_entropies_and_entropic_scale_parametrization}d illustrates that a high entropic scale \mbox{($E_s\!\!=\!10$)} produces even higher entropy probability distributions and the highest OOD detection performance regardless of the out-distribution considered (Fig.~\ref{fig:train_losses_entropies_and_entropic_scale_parametrization}e). 

Despite the entropy scale value used, the cross-entropy is minimized and the classification accuracies produced by SoftMax and IsoMax losses are similar. \emph{For a high entropic scale, the IsoMax loss minimizes the cross-entropy while producing high entropy posterior probability distributions as recommended by the principle of maximum entropy}.

More importantly, higher entropy posterior probability distributions directly correlate with increased OOD detection performances despite the OOD data. \emph{Fig.~\ref{fig:train_losses_entropies_and_entropic_scale_parametrization}d shows that an entropic scale \mbox{$E_s\!\!=\!10$} is enough to produce probability distributions with essentially the maximum possible entropies. Hence, no additional gains may be obtained by increasing the entropic scale even further.} Indeed, we experimentally observed no further gains for entropic scales higher than ten.

%\emph{After defining \mbox{$E_s\!\!=\!10$} for the IsoMax loss based on the previous experiments}, we performed additional analyses \emph{using validation data}.
Fig.~\ref{fig:entropic_score_study} shows that \emph{regardless of the combination of dataset and model}, the classification accuracy and the mean OOD detection performance are stable for \mbox{$E_s\!\!=\!10$} or higher, as the entropic scale is already high enough to ensure near-maximal entropy. We speculate that \mbox{$E_s\!\!=\!10$} worked satisfactorily regardless of the number of training classes because it is used inside an exponential function while the entropy increases only with the logarithm of the number of training classes. Hence, an eventual validation of $E_s$ would produce an insignificant performance increase. Actually, this is not possible because we consider access to OOD or outlier samples forbidden. Moreover, making $E_s$ learnable did not significantly affect the OOD detection performance.

In other words, considering that the entropic scale $E_{s}$ is present in Equation \eqref{eq:loss_isomax}, we are initially led to think that it needs to be tuned to achieve the highest possible OOD detection performance. However, we emphasize that the experiments showed that the dependence of the OOD detection performance on the entropic scale is remarkably well-behaved. \emph{Essentially, the OOD detection performance monotonically increases with the entropic scale until it reaches a saturation point near $E_{s}\!\!=\!10$ regardless of the dataset or the number of training classes (Fig.~\ref{fig:entropic_score_study})}. It may be explained by the fact that the entropic scale is inside an exponential function and that we experimentally observed that $\exp{(-10\!\times\!d)}$ is enough to produce almost maximum entropy regardless of the dataset, model or number of training classes under consideration. Finally, even if we consider that validating for entropic scales higher than ten would allow some minor {\color{black}OOD} detection performance improvement, we usually cannot do this because we do not often have access to out-distribution data.

Hence, the well-behaved dependence of the OOD detection performance on the entropic scale allowed us to define the $E_{s}$ as a \emph{constant scalar} equals to 10 rather than a hyperparameter that needs to be tuned for each novel dataset and model. It is the reason our approach may be used without requiring access to OOD/outlier data. Therefore, \emph{we kept the entropic scale as a constant (global value) equal to ten for all subsequent experiments}. Hence, no validation of the entropic score was performed for different models, datasets, or number of classes.

\subsection{Fair Comparison: SoftMax versus IsoMax}

Table~\ref{tbl:expanded_fair_odd} shows that models trained with the SoftMax loss using the maximum probability as the score \mbox{(SoftMax+MPS)} almost always present the worst OOD detection performance.

In the case of models trained with SoftMax loss, replacing the maximum probability score by the entropic score (SoftMax+ES) produces small OOD detection performance gains. However, the combination of models trained using IsoMax loss with the entropic score (IsoMax+ES), which is the proposed solution, significantly improves, usually by several percentage points, the OOD detection performance across almost all datasets, models, out-distributions, and~metrics. \emph{We emphasize that the entropic score only produces high OOD detection performance when the probability distributions present high entropy (IsoMax+ES). For low entropy probability distributions (SoftMax+ES), the performance increase is minimal}. For a study showing that the IsoMax loss superiority in relation to SoftMax is robust for a different number of training examples per class, see Supplementary Material \ref{apx:robustness_study}.

The IsoMax loss is well-positioned to replace SoftMax loss as a novel baseline OOD detection approach for neural networks OOD detection, as the former does not present an accuracy drop in relation to the latter and simultaneously improves the OOD detection performance. Additional techniques (e.g., input preprocessing, adversarial training, and outlier exposure) may be added to improve OOD detection performance gains further.

%\subsection{Classification Accuracy}
Table \ref{tbl:addon_odd} shows that IsoMax loss consistently produces classification accuracy similar to SoftMax loss, regardless of being used as baseline approach or combined with additional techniques to improve OOD detection performance. The Supplementary Material \ref{apx:robustness_study} presents further evidence that IsoMax loss does not present classification accuracy drop in comparison with SoftMax loss.

The IsoMax loss enhanced versions almost always outperform the OOD detection performance of the corresponding SoftMax loss enhanced version when both are using the same add-on technique. The major drawback of adding label smoothing or center loss regularization is the need to tune the hyperparameters presented by these add-on techniques. Additionally, different hyperparameters values need to be validated for each pair of in-distribution and out-distribution. As mentioned before, it is highly optimistic to assume access to OOD data, as we usually do not know what OOD data the solution we will face in the field. \emph{Even considering this best-case scenario, SoftMax loss combined with label smoothing or center loss regularization always presented significantly lower OOD detection performance than IsoMax loss without using them and, consequently, avoiding unrealistically optimist access to OOD data and the mentioned validations.} Enhancing SoftMax loss or IsoMax loss with ODIN presents the same problems from a practical perspective. \emph{In CIFAR100, SoftMax loss with outlier exposure produces lower performance than IsoMax loss without it for both DenseNet and ResNet.}

\textcolor{black}{The Table~\ref{tbl:text_odd} presents results for text data. As expected, the results show that our approach is \emph{domain-agnostic} and therefore may be applied to data other than images.}

%\subsubsection{Unfair Scenario}
\subsection{Unfair Comparison: SoftMax versus No Seamless Solutions}

Table~\ref{tbl:unfair_odd} presents a perspective about how our proposed baseline OOD detection approach compares with no seamless solutions. Hence, we need to analyze the mentioned table considering that it shows an \emph{unfair} comparison of approaches that present different requirements and side effects. ODIN and the Mahalanobis solution use input preprocessing\footnote{To allow OOD detection, each inference requires a first neural network forward pass, a backpropagation, and a second forward pass.}.

Consequently, they present solutions with much slower and less energy-efficient inferences than models trained with IsoMax loss, which are as fast and computation-efficient as the models trained with common SoftMax loss. Moreover, they also require validation using adversarial samples.

Additionally, ODIN requires temperature calibration, while the Mahalanobis approach uses feature ensemble and metric learning, which may be implicated in limited scalability for large-size images. ACET requires adversarial training, which may also prevent its use in large-size images. ODIN, the Mahalanobis approach, and ACET have hyperparameters tuned for each combination of in-data and models in the table. ODIN, the Mahalanobis method, and ACET used SoftMax loss rather than IsoMax loss. IsoMax+ES exhibits neither the mentioned unfortunate requirements nor undesired side effects. 

%Finally, we emphasize that the adversarial hyperparameters (e.g., the maximum adversarial perturbation) used in ODIN/Mahalanobis/ACET methods were validated in previous papers not related to OOD detection using the SVHN/CIFAR10/CIFAR100 validation sets. However, these validations sets were also used in the competing OOD methods papers as {\color{blue}OOD} detection test sets. Hence, the {\color{blue}OOD} detection performances reported in the compared methods papers used here for comparison may be overestimated.

{\color{black}Additionally, taking into consideration that ODIN, Mahalanobis, and ACET used adversarial hyperparameters (e.g., the adversarial perturbation) that were validated using the validation sets of SVHN/CIFAR10/CIFAR100 and noticing that these sets composed the OOD detection test sets, we conclude that some overestimation may be present in the {\color{black}OOD} detection performance reported by these papers.}

Regardless of the previous considerations, Table~\ref{tbl:unfair_odd} shows that IsoMax+ES considerably outperforms ODIN in all evaluated scenarios. Therefore, in addition to avoiding hyperparameter tuning and access to OOD or adversarial samples, the results show that removing the entropic scale is much more effective in increasing the OOD detection performance than performing temperature calibration.

Furthermore, IsoMax+ES usually outperforms ACET by a large margin. Moreover, in most cases, even operating under much more favorable conditions, the Mahalanobis method surpasses IsoMax+ES by less than 2\%. In some scenarios, the latter overcomes the former despite avoiding hyperparameter validation, being seamless and producing much faster and more computation-efficient inferences, as no input preprocessing technique is~required.

\subsection{Inference Efficiency}

Table~\ref{tbl:times} presents the inference delays for SoftMax loss, IsoMax loss, and competing methods using CPU and GPU. We observe that neural networks trained using IsoMax loss produce inferences equally fast as the ones produced by networks trained using SoftMax loss, regardless of using CPU or GPU for inference. Additionally, the entropic score is as fast as the usual maximum probability score. Moreover, methods based on input preprocessing were more than ten times slower on CPU and about four times slower on GPU. Finally, those rates presumably apply to the computational cost and energy consumption as well.

At first sight, inference methods (i.e., methods that can be applied to pre-trained models) may be seen as ``low cost'' compared with training methods like our IsoMax loss, as we avoid training or fine-tuning the neural network. However, this conclusion may be misleading, as we have to keep in mind that inference methods (e.g., ODIN~\cite{liang2018enhancing} and Mahalanobis~\cite{lee2018simple}) produce inferences that are much more energy, computation, and time inefficient (Table~\ref{tbl:times}). For example, consider initially the rare practical situation where a pretrained model is available, and no fine-tuning to a custom dataset is required. In such cases, an inference method may indeed be applied without requiring any loss function. However, despite avoiding training or fine-tuning a neural network once or a few times, all the subsequent inferences, which are usually performed thousands or millions of times on the field (sometimes even by constrained devices), will be about 6 to 10 times more computational, energy, environment, and time inefficient (Table~\ref{tbl:times}).

Alternatively, consider the case where a pretrained model is not available or fine-tuning to a custom dataset is needed. In this situation, which is much more likely in practice, we cannot avoid training or fine-tuning the neural network, and a training method like ours will be required anyway. In these cases, it would be recommended to train or fine-tune using IsoMax rather than SoftMax loss, as our experiments showed that both training times are the same and IsoMax loss produces considerably higher OOD detection performance.

Hence, inference methods are more inference inefficient because they coexist with a model that was trained with a loss not designed from the start with OOD detection in mind. The drawback is to produce an enormous amount of inefficient inferences on the field that usually uses constrained computational resource devices such as embedded systems.

Nevertheless, suppose the increased inference time, computation, environment damage, and energy consumption required to use an inference method is not a concern from a practical point of view. In that case, the model pretrained or fine-tuned using a training method may be subsequently subjected to the desired inference approach to increase overall OOD detection performance further. In other words, we do not claim that inference methods such as ODIN and Mahalanobis do not increase the OOD detection performance compared with Maximum Score Probability or even the Entropic Score. However, we point out that producing more energy inefficient inferences is one drawback of adopting such inference methods.

In summary, rather than concurrent, the training and inference methods are orthogonal and complementary. Moreover, we see no reason not to train the models with a loss designed to support OOD, regardless of subsequently applying an OOD inference method.

\subsection{Additional Studies}

For additional analyses regarding logits, probabilities, and entropies, please see Supplementary Material~\ref{apx:additional_analyses}. For information regarding the similarity of the SoftMax loss and the IsoMax loss training metrics, see the Supplementary Material~\ref{apx:training_metrics}.

\section{Conclusion}\label{sec:conclusion}

We proposed a baseline OOD detection approach based on the maximum entropy principle. The proposed IsoMax loss works as a SoftMax loss drop-in replacement that produces accurate predictions in addition to inferences that are fast and energy- and computation-efficient. OOD detection is performed using the rapid entropic score. Furthermore, it is also turnkey, as no additional procedures other than a straightforward neural network training is required, and no hyperparameters are tuned. Neither is required collection of extra/outlier data.

The direct replacement of the SoftMax loss by the IsoMax loss significantly improves the baseline neural networks' OOD detection performance. Hence, we conclude that the general low OOD detection performance of current deep networks is due to SoftMax loss drawbacks, i.e., anisotropy and overconfidence, rather than the models' limitations.

Our OOD detection approach produces high OOD detection performance without relying on extra techniques (e.g., input preprocessing, adversarial training, hyperparameter validation, feature ensemble, additional/outlier/background data, and others) that, unfortunately, present inconvenient requirements and undesired side effects. Nevertheless, when the cited requirements and side effects are not a concern for a particular real-world application, the mentioned (or novel) techniques may be combined with IsoMax loss to possibly achieve higher OOD detection performance. This combined solution may be particularly needed for more challenging large-size image datasets such as ImageNet. 

Summarily, the three most original and relevant contributions of this paper are the following: First, the theoretical arguments, intuitions, and insights that motivate neural network design with high entropy outputs. Second, ``the entropy maximization trick'' that allows it. Third, the experimental demonstration that the entropy of neural networks with high entropy posterior probability distributions represents a high-performance OOD detection score. 

We intend to apply our OOD detection approach to models using large-size~images used in large-scale real-world applications in future works, as we believe our approach scales to these cases as well. %{\color{blue}We believe more tests are need to evaluate the behavior of the entropic scale in datasets with thousands of classes.}
As the principle of maximum entropy is a general concept, we also plan to adapt our approach to machine learning areas other than deep learning. Another promising approach could be using recent data augmentation techniques \cite{NIPS2019_9540, yun2019cutmix} or strategies based on pretrained models \cite{Sastry2019DetectingOE}. Additionally, despite requiring auxiliary/outlier/background data, to further increase the OOD detection performance of our solution, we plan to incorporate loss enhancement (regularization) techniques such as outlier exposure \cite{hendrycks2018deep,papadopoulos2019outlier}, background samples \cite{NIPS2018_8129} and energy-based training and score techniques \cite{DBLP:journals/corr/abs-2010-03759}.

%\newpage
\bibliographystyle{IEEEtran}
%\interlinepenalty=10000
\bibliography{references}
%\bibliography{Mendeley.bib, new_references.bib}{}

%\begin{comment}
\begin{IEEEbiography}[{\includegraphics[width=1in, height=1.25in, clip,keepaspectratio]{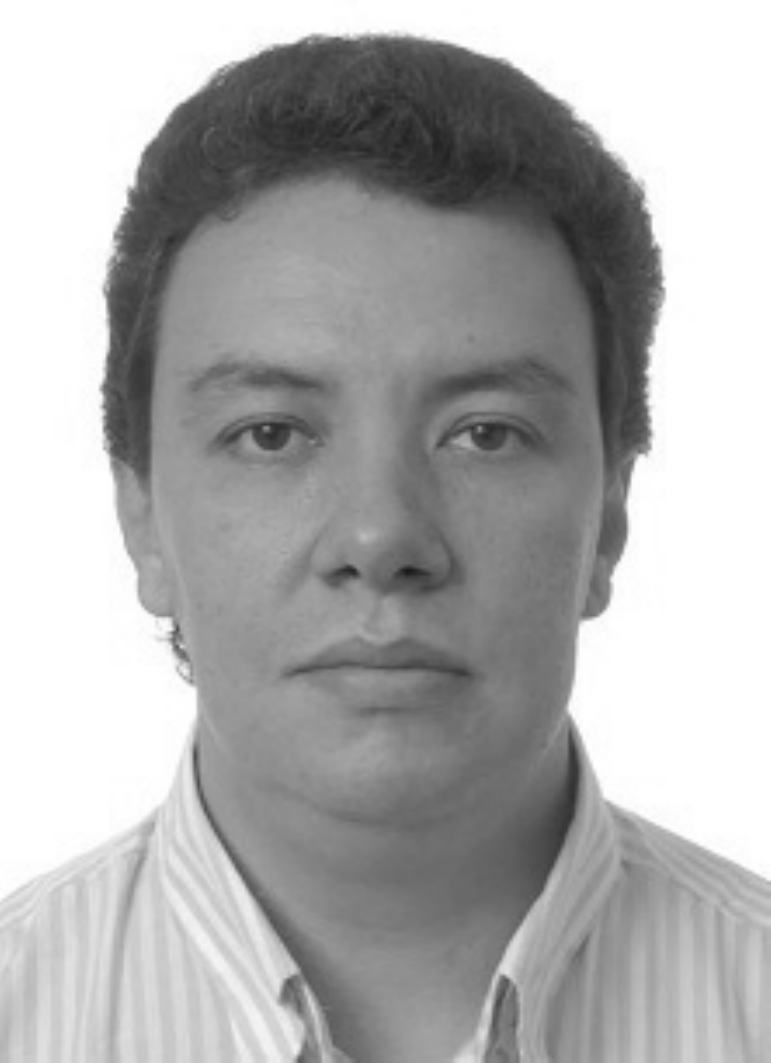}}]{David Mac\^edo}
%\begin{IEEEbiography}{Michael Shell}
received his M.Sc. Computer Science degree and B.Sc. Electronic Engineer degree summa cum laude from the Universidade Federal de Pernambuco (UFPE), Brazil. He was a visiting researcher at the Montreal Institute for Learning Algorithms (MILA), Université de Montréal (UdeM), Canada. He co-created and is currently a collaborator professor of the Deep Learning course of the Computer Science Master and Doctorate Programs at the Center for Informatics (CIn), Universidade Federal de Pernambuco (UFPE), Brazil. He is a professor at Faculdade Nova Roma, Brazil. He has authored dozens of deep learning papers published in international peer-reviewed journals and conferences and is a reviewer of IEEE journals. He is pursuing a Ph.D. degree in computer science at UFPE. His research interests include Deep Learning, Trustworthy Artificial Intelligence, Computer Vision, Natural Language Processing, and Speech Processing.
\end{IEEEbiography}

\begin{IEEEbiography}[{\includegraphics[width=1in, height=1.25in, clip,keepaspectratio]{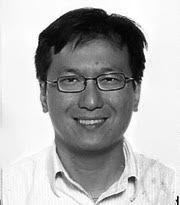}}]{Tsang Ing Ren}
%%%\begin{IEEEbiography}{Michael Shell}
received the B.Sc. degree in electronic engineering from the Universidade Federal de Pernambuco, Recife, Brazil, and the Ph.D. degree in physics from the University of Antwerp, Antwerp, Belgium.  He is currently an associate professor with the Center for Informatics, Universidade Federal de Pernambuco. His research interests include Machine Learning, Image Processing, Computational Photography, Computer Vision, and Deep Learning.
\end{IEEEbiography}

\begin{IEEEbiography}[{\includegraphics[width=1in, height=1.25in, clip,keepaspectratio]{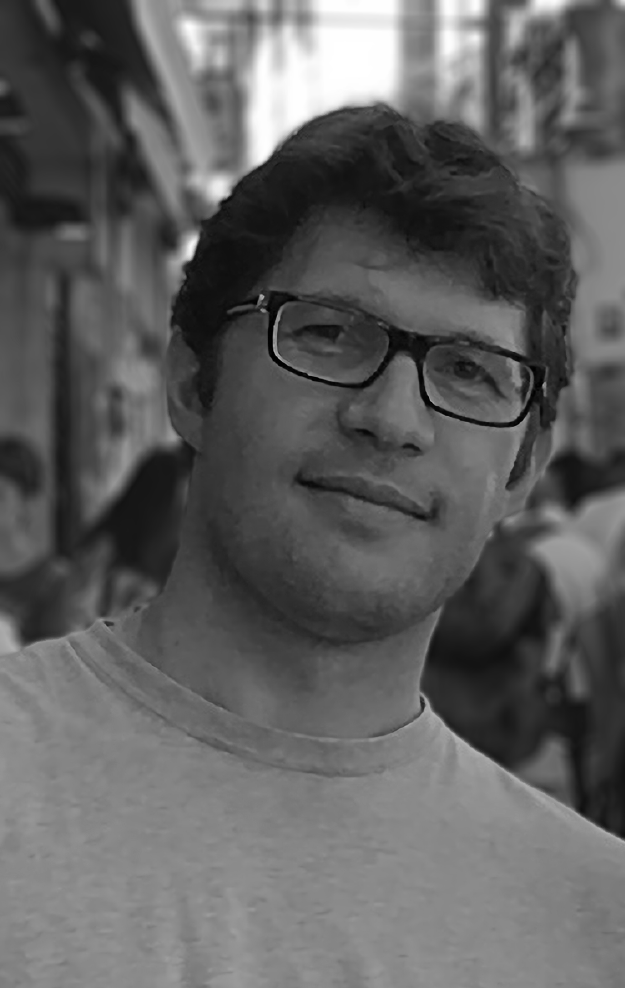}}]{Cleber Zanchettin}
%%%\begin{IEEEbiography}{Michael Shell}
received his Ph.D. degree in computer science from the Universidade Federal de Pernambuco, Recife, Brazil, in 2008. He is currently a professor and a technical reviewer with the Centro de Inform\'atica, Universidade Federal de Pernambuco. He has authored over 60 papers in international refereed journals and conferences in pattern recognition, artificial neural networks (ANNs), and intelligent systems. His current research interests include hybrid neural systems and applications of ANNs.
\end{IEEEbiography}

\begin{IEEEbiography}[{\includegraphics[width=1in, height=1.25in, clip,keepaspectratio]{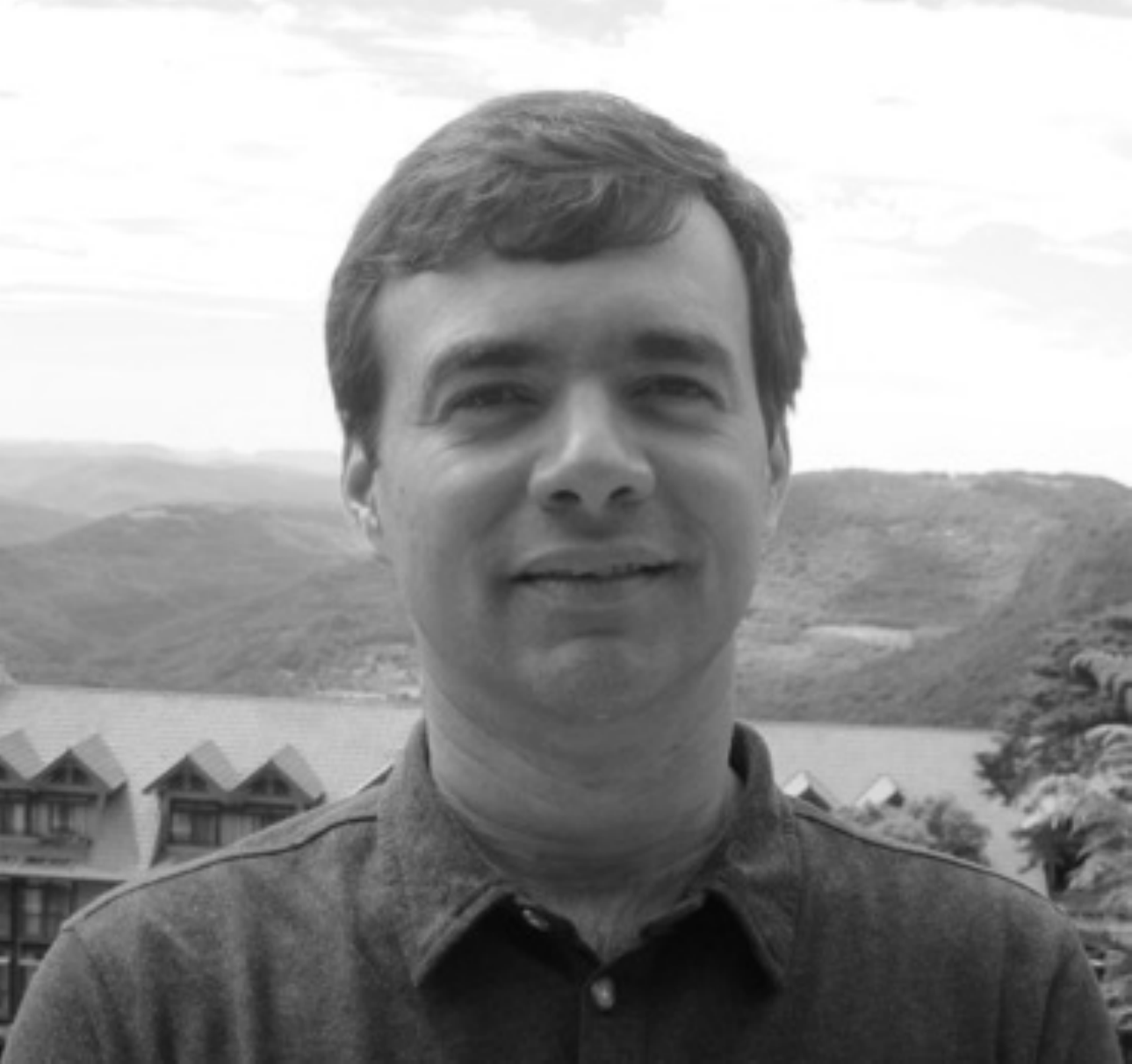}}]{Adriano L. I. Oliveira}
%%%\begin{IEEEbiography}{Michael Shell}
obtained his B.Sc. degree in electrical engineering and M.Sc. and Ph.D. degrees in computer science from the Universidade Federal de Pernambuco, Brazil, in 1993, 1997 and 2004, respectively. In 2011, he joined the Center for Informatics at the Universidade Federal de Pernambuco as an assistant professor. He was a visiting professor at École de technologie supérieure, Université du Quebec, Montreal, Canadá, from 2018 to 2019. He was an assistant professor at the Universidade Federal Rural de Pernambuco from 2009 to 2011 and at the Pernambuco State University from 2002 to 2009. He has published over 110 articles in scientific journals and conferences and one book. He is a senior member of the IEEE. His current research interests include neural networks, machine learning, pattern recognition, data mining, and applications of these techniques to time series analysis and forecasting, information systems, software engineering, and biomedicine.
\end{IEEEbiography}

\begin{IEEEbiography}[{\includegraphics[width=1in, height=1.25in, clip,keepaspectratio]{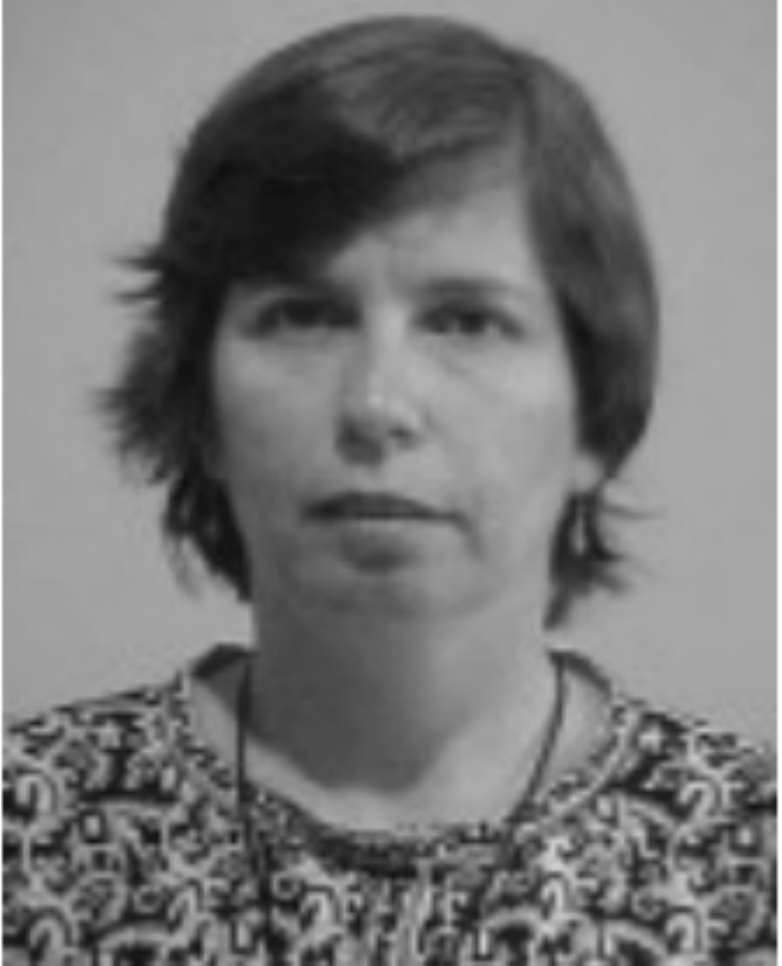}}]{Teresa Ludermir}
%%%\begin{IEEEbiography}{Michael Shell}
received her Ph.D. degree in artificial neural networks from the Imperial College London, London, U.K., in 1990. She was a lecturer with Kings College London, London, from 1991 to 1992. She joined the Centro de Inform\'atica, Universidade Federal de Pernambuco, Recife, Brazil, in 1992, where she is currently a professor and the Head of the Computational Intelligence Group. She has authored over 300 articles in scientific journals and conferences and three books on neural networks. Her current research interests include weightless neural networks, hybrid neural systems, and applications of neural networks. Ludermir organized two of the Brazilian Symposiums on Neural Networks.
\end{IEEEbiography}
%\end{comment}

%\newpage
%
%\appendices
%\section{Proof of the First Zonklar Equation}
%Appendix one text goes here.

\clearpage
\begin{center}{\bf {\LARGE Supplementary Material}}\end{center}
\appendices
%\appendix
%\onecolumn
%\twocolumn

\begin{figure*}%[!h]
%\vskip -0.25cm
\centering
%\subfloat[]{\includegraphics[width=\textwidth]{plots/plot_odd1_robustness_acc1_densenetbc100.png}}
%\\
%\vskip -0.1cm
\subfloat[]{\includegraphics[width=0.9\textwidth]{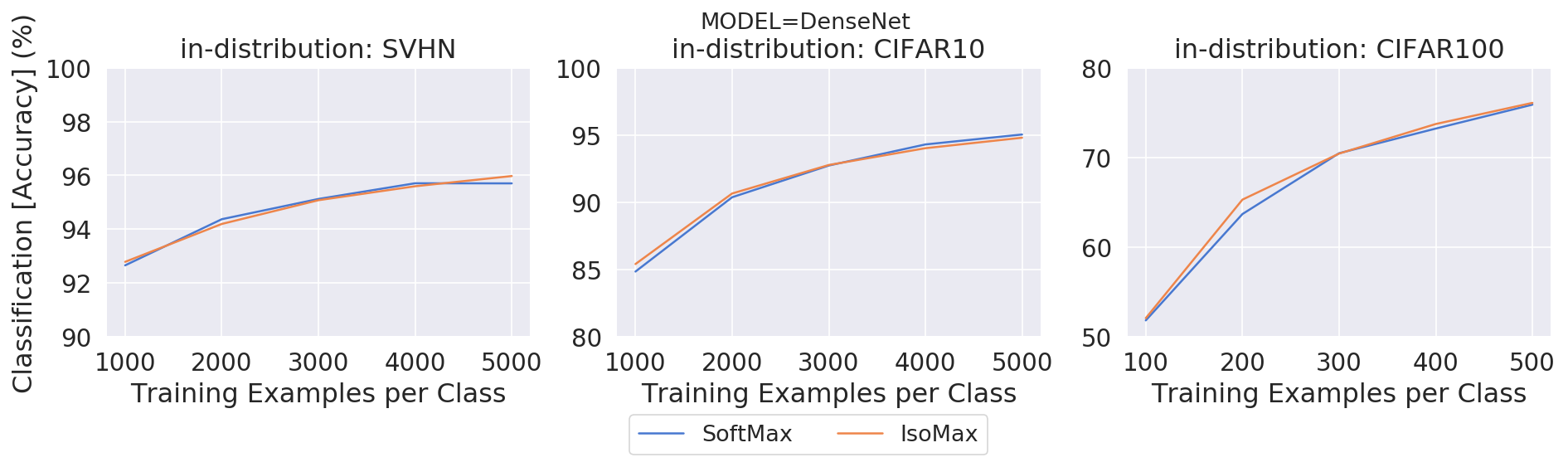}} 
\\
%\vskip 0.05cm
\subfloat[]{\includegraphics[width=0.9\textwidth]{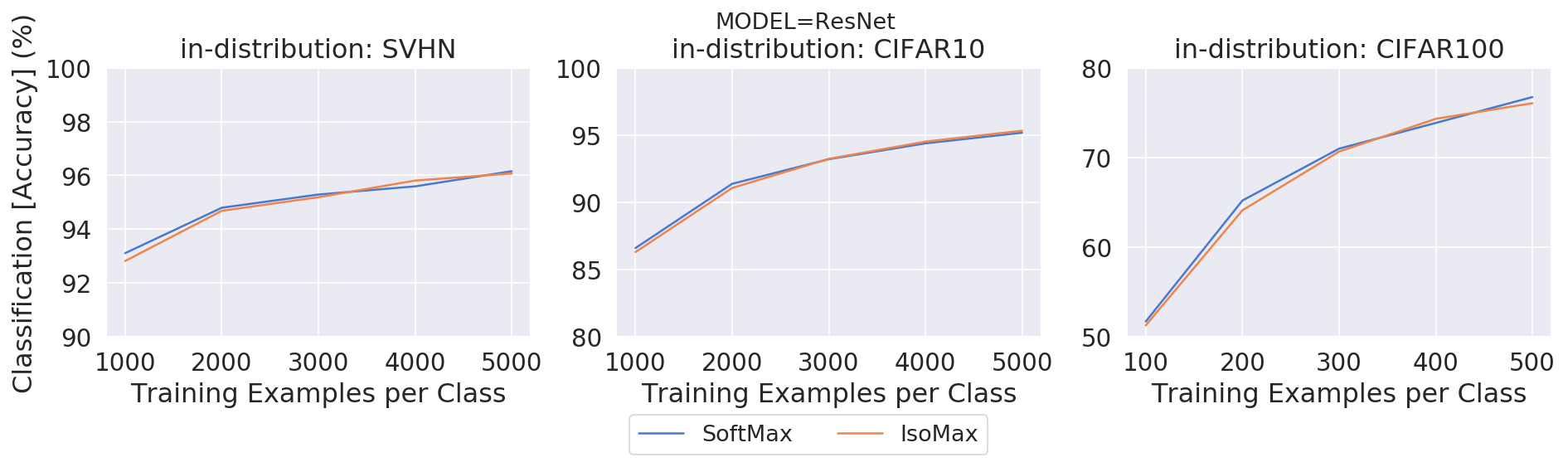}}
%\vskip 0.05cm
\caption{IsoMax loss presents test accuracy similar to SoftMax loss (no classification accuracy drop) for different numbers of training examples per class on several datasets and models. Simultaneously, IsoMax usually produces a much higher OOD detection performance (Table~\ref{tbl:expanded_fair_odd}).}
\label{fig:classification_robstness}
\end{figure*}

\begin{figure*}%[!h]
%\vskip -0.25cm
\centering
%\subfloat[]{\includegraphics[width=\textwidth]{plots/plot_odd1_robustness_acc1_densenetbc100.png}}
%\\
%\vskip -0.1cm
\subfloat[]{\includegraphics[width=0.95\textwidth]{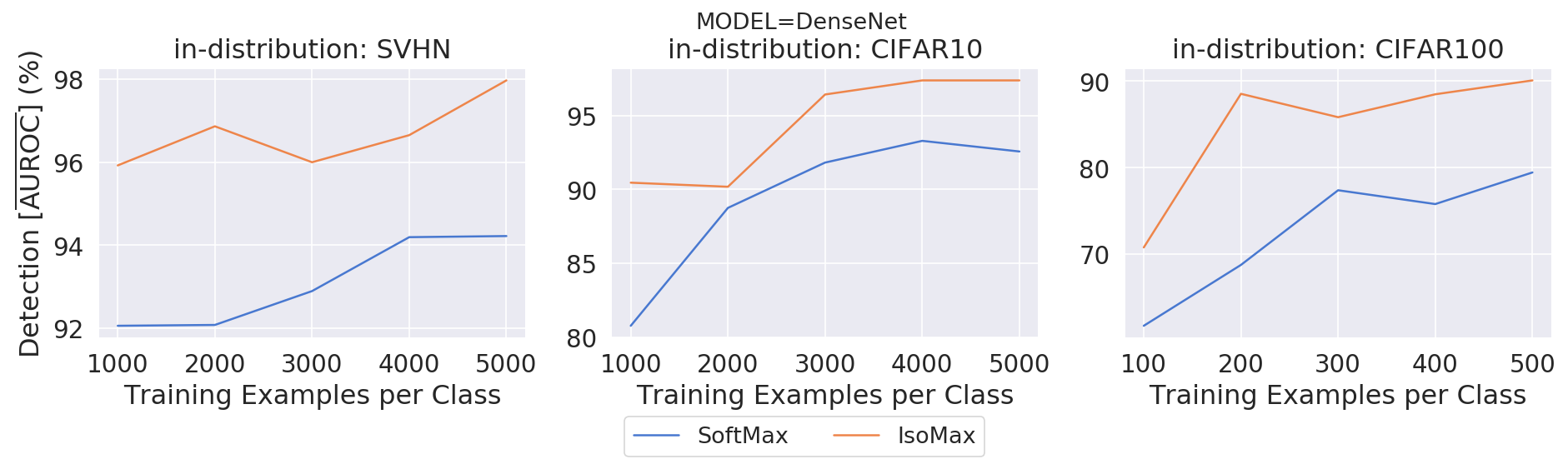}} 
\\
\vskip -0.05cm
\subfloat[]{\includegraphics[width=0.95\textwidth]{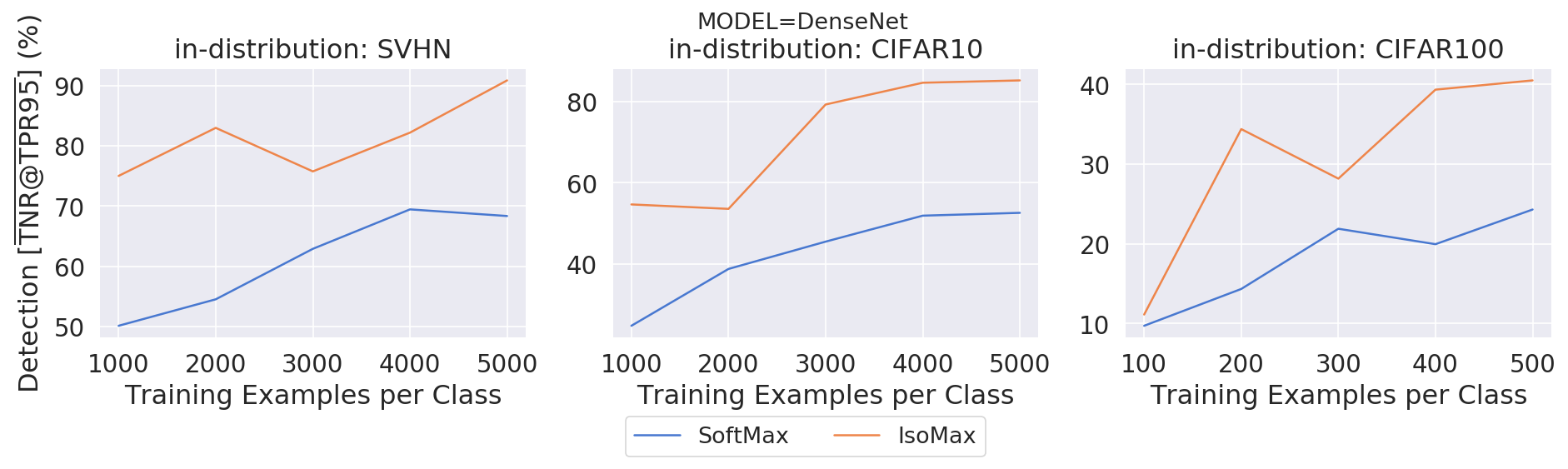}}
\\
\vskip -0.05cm
\subfloat[]{\includegraphics[width=0.95\textwidth]{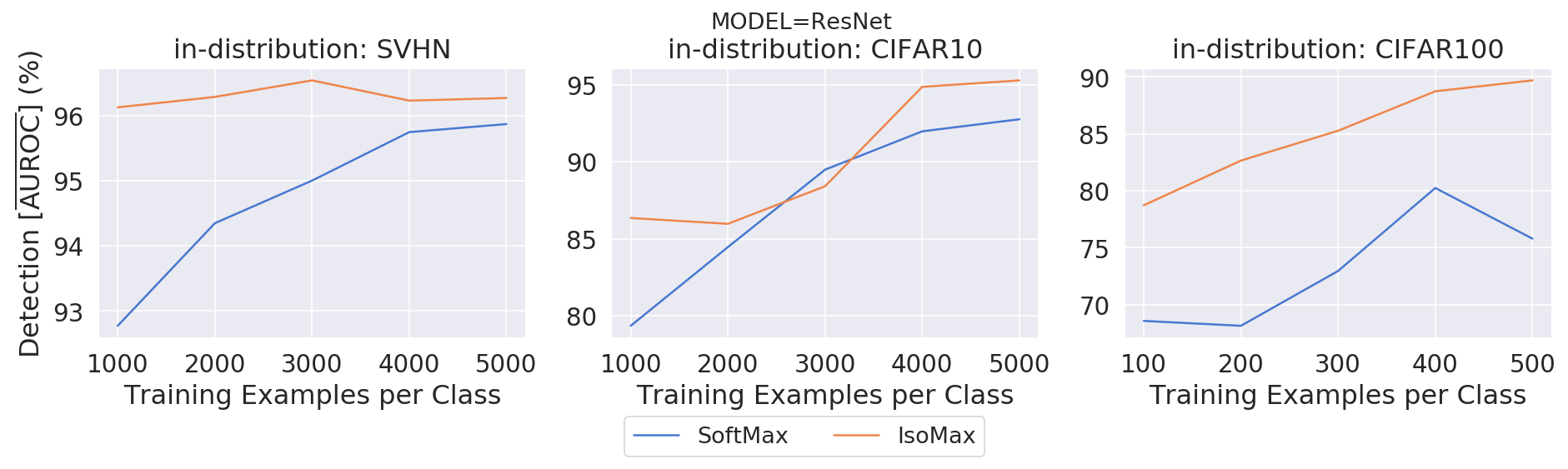}} 
\\
\vskip -0.05cm
\subfloat[]{\includegraphics[width=0.95\textwidth]{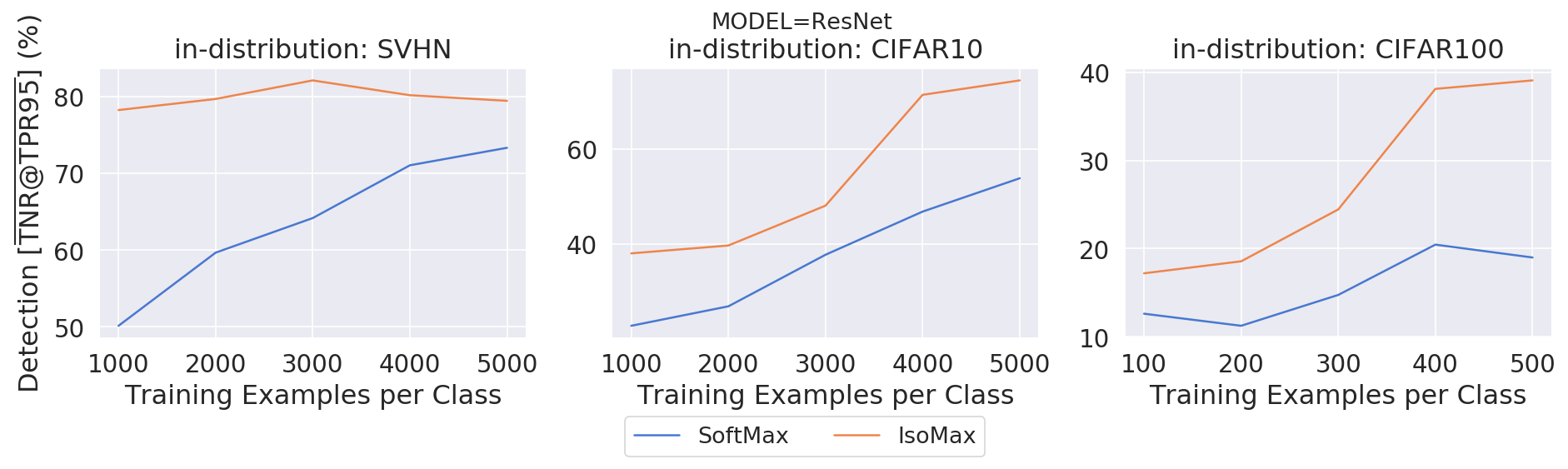}}
\vskip -0.05cm
%\caption{Robustness study: IsoMax loss consistently presents higher OOD detection performance than does SoftMax loss for different numbers of training examples per class on several datasets, models, and metrics. The entropic score was used for both SoftMax and IsoMax losses. For each in-distribution, we calculated the mean AUROC and TNR@TPR95 considering all possible OOD detection test sets{\color{red}, except noise and fooling ones} (see Section~\ref{sec:experiments}~for~extra~details).}
\caption{Robustness study: IsoMax loss consistently presents higher OOD detection performance than does SoftMax loss for different numbers of training examples per class on several datasets, models, and metrics. The entropic score was used for both SoftMax and IsoMax losses. For each in-distribution, we calculated the mean AUROC and TNR@TPR95 considering all possible OOD detection test sets~for~extra~details).}
\label{fig:detection_robstness}
\end{figure*}

\begin{figure*}%[t]
%\vskip -0.25cm
\centering
\subfloat[]{\includegraphics[width=0.95\textwidth]{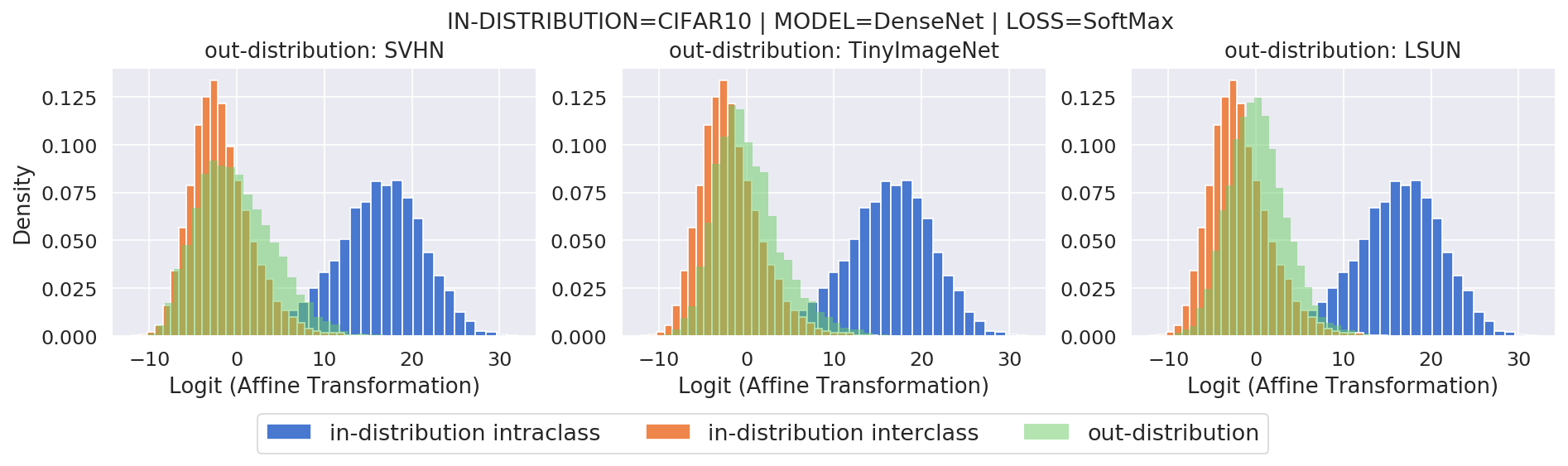}\label{fig:softmax_logits_histograms}}
\\
%\vskip -0.1cm
\subfloat[]{\includegraphics[width=0.95\textwidth]{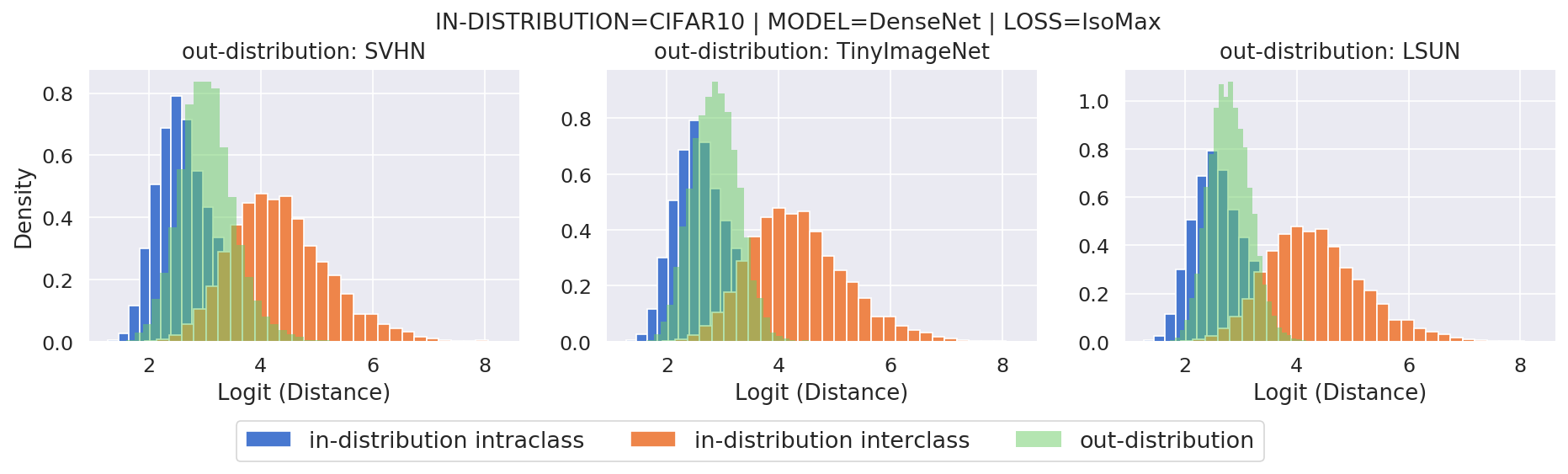}\label{fig:isomax2_logits_histograms}} 
%\\
\vskip -0.15cm
\caption{Logits: (a) In SoftMax loss, out-distribution logits mimic in-distribution interclass logits. (b) In IsoMax loss, out-distribution logits mimic in-distribution intraclass logits, which facilitates OOD detection, as there are many more interclass logits than intraclass logits, and the entropic score takes into consideration the information provided by all network outputs rather than just one. Distances are calculated from class prototypes.}
\label{fig:logits_histograms}
\end{figure*}

\begin{figure*}%[!h]
\centering
\subfloat[]{\includegraphics[width=0.925\textwidth]{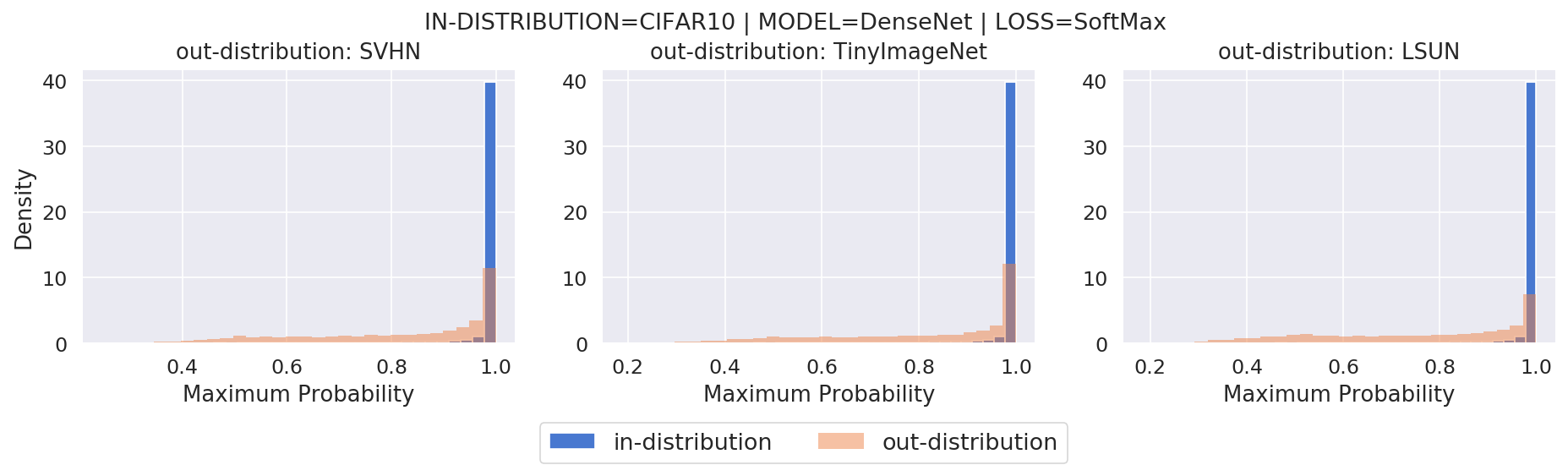}\label{fig:softmax_maxprobs_histograms}}
\\
\vskip -0.05cm
\subfloat[]{\includegraphics[width=0.925\textwidth]{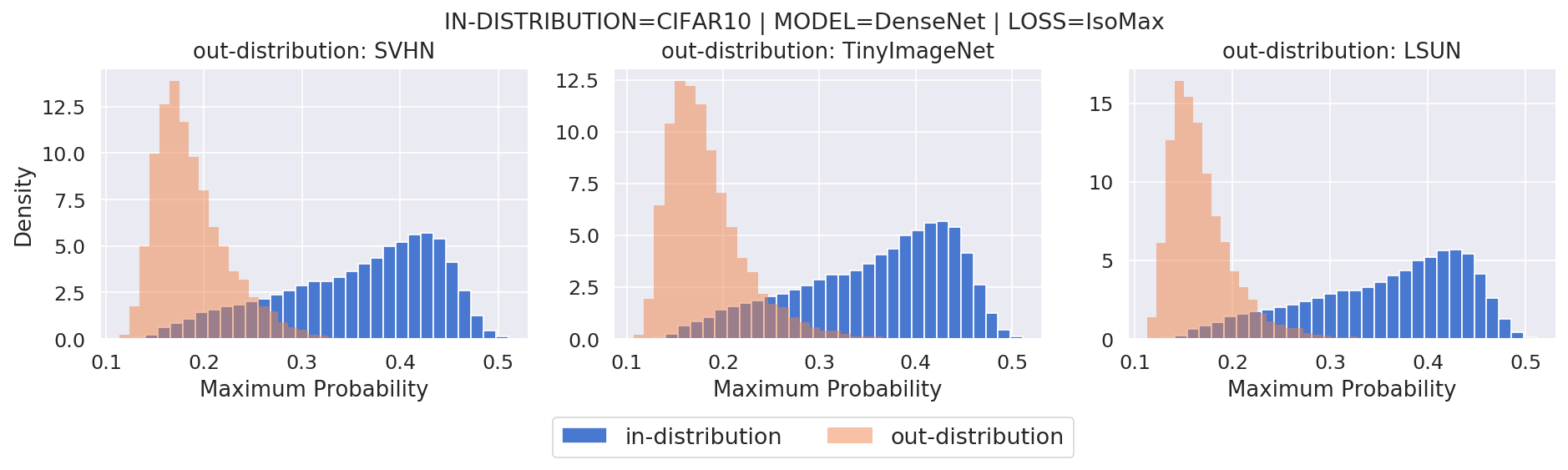}\label{fig:isomax_maxprobs_histograms}} 
\\
\vskip -0.05cm
\subfloat[]{\includegraphics[width=0.925\textwidth]{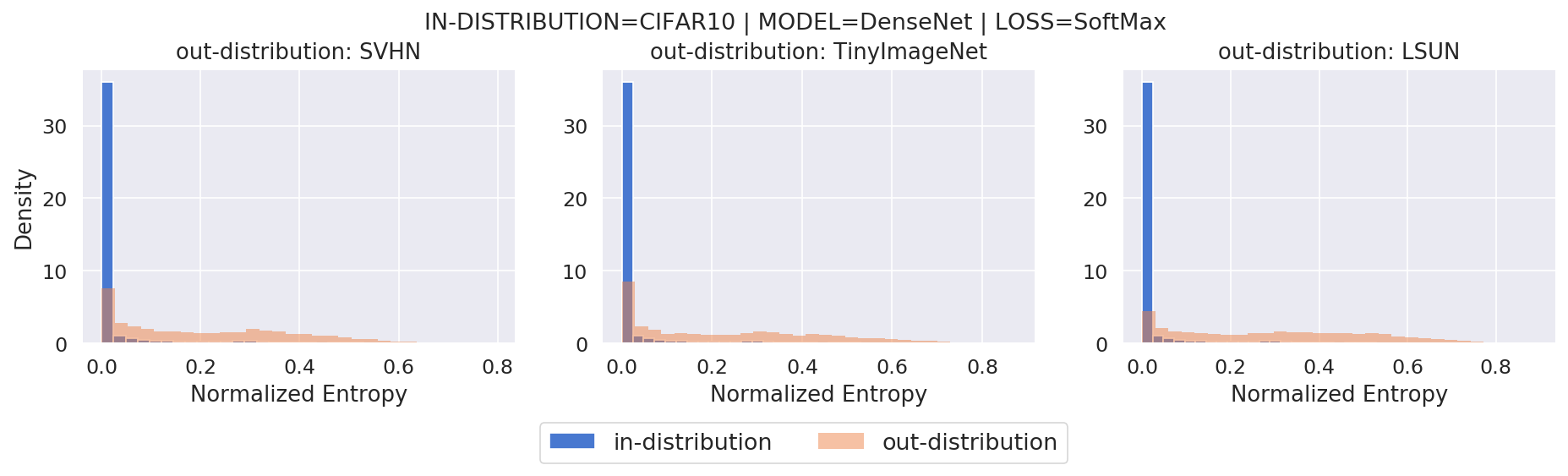}\label{fig:softmax_entropies_histograms}}
\\
\vskip -0.05cm
\subfloat[]{\includegraphics[width=0.925\textwidth]{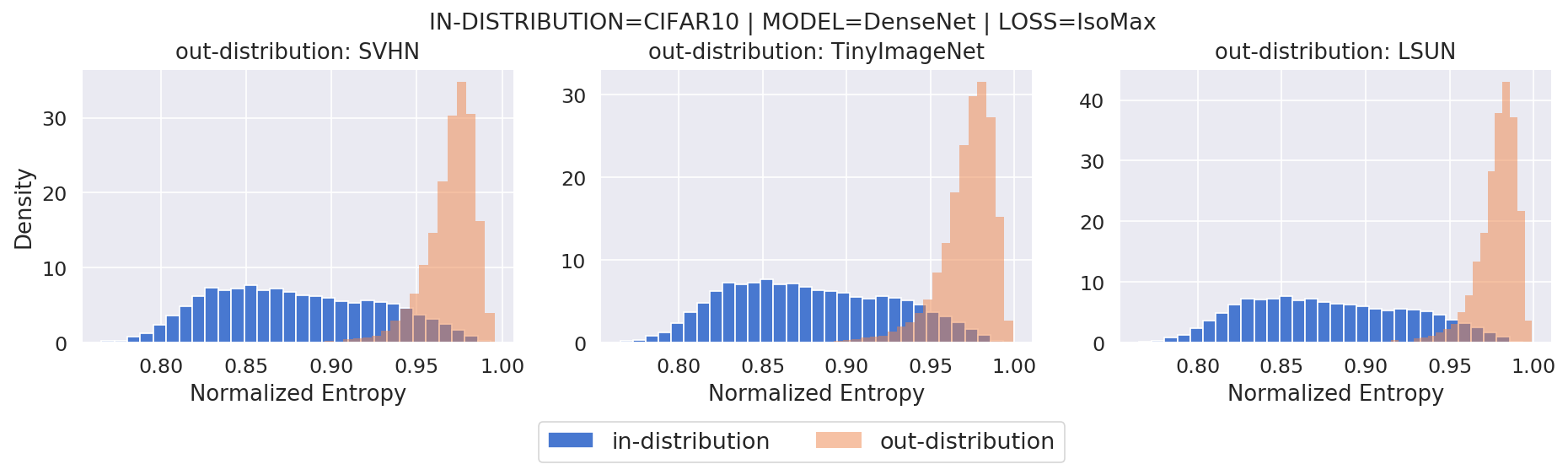}\label{fig:isomax_entropies_histograms}} 
\vskip -0.05cm
\caption{Probabilities and entropies: (a) SoftMax loss produces high confident predictions for in-distribution samples. SoftMax loss usually provides high maximum probabilities even for OOD samples. (b) IsoMax loss produces less confident predictions for in-distribution samples than SoftMax loss. IsoMax loss commonly produces even lower maximum probability for OOD samples. (c) SoftMax loss provides low entropy (high confidence) for almost all in-distribution samples and even usually for OOD samples. (d) IsoMax loss produces high entropy for out-distributions. More precise separation between the in-distribution and out-distributions is obtained. \emph{Following the principle of maximum entropy, rather than calibrated probabilities, we need to provide predictions as underconfident (high entropy) as possible as long as they fit the data appropriately, i.e., produce no classification accuracy drop}.}
\label{fig:maxprobs_entropies_histograms}
\end{figure*}

\begin{figure*}%[!t]
\centering
\subfloat[]{\includegraphics[width=\textwidth]{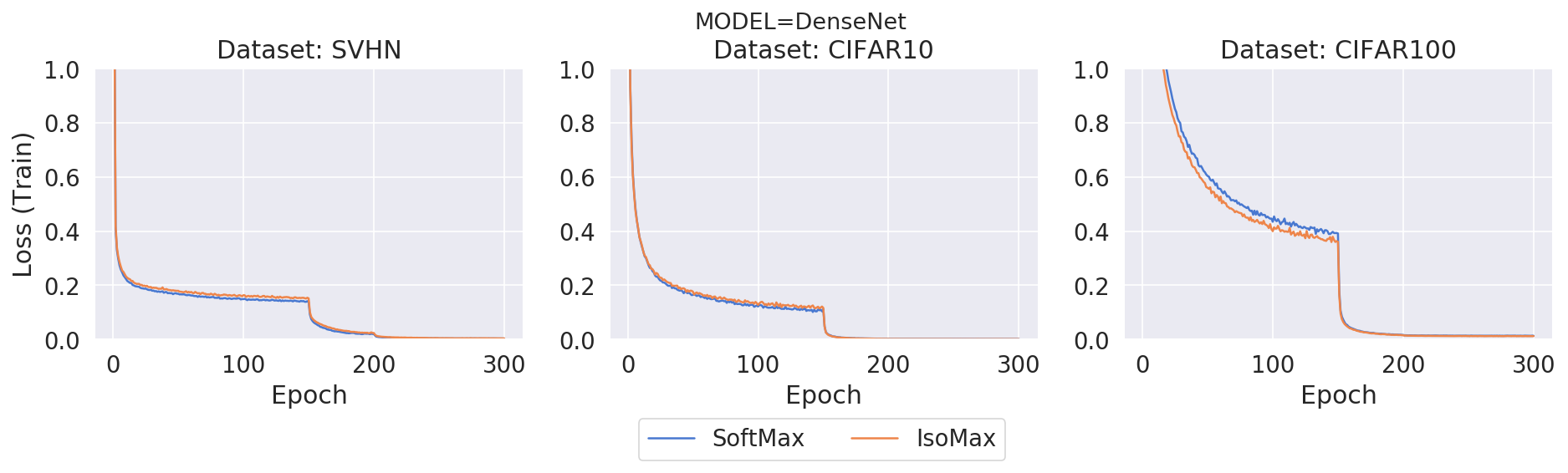}}
\\
\vskip -0.025cm
\subfloat[]{\includegraphics[width=\textwidth]{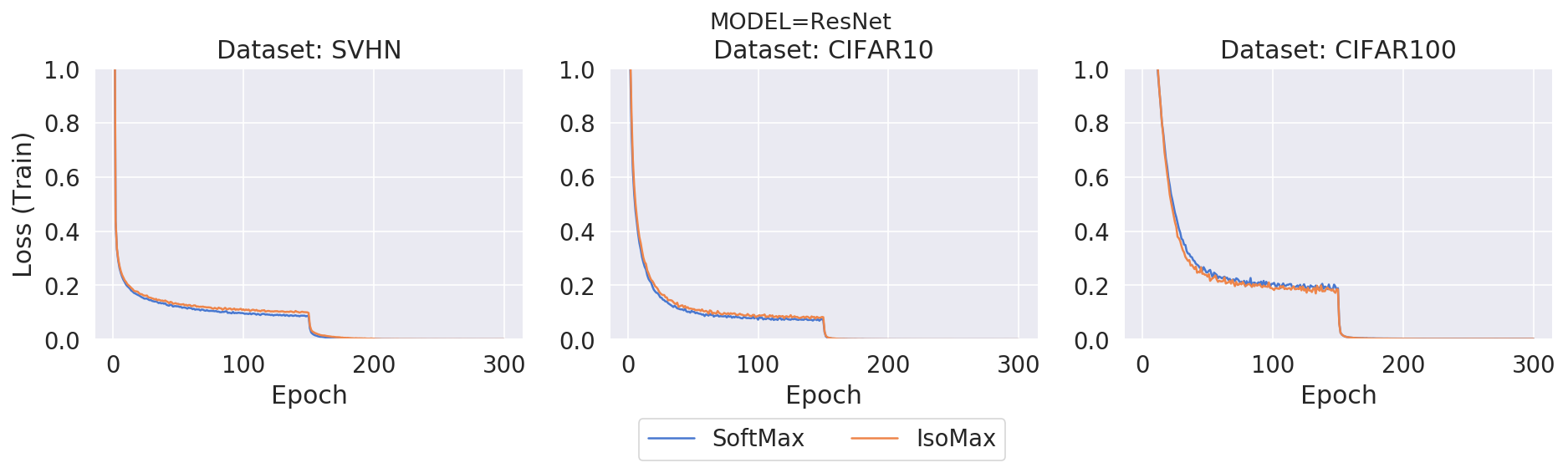}}
\\
\vskip -0.025cm
\subfloat[]{\includegraphics[width=\textwidth]{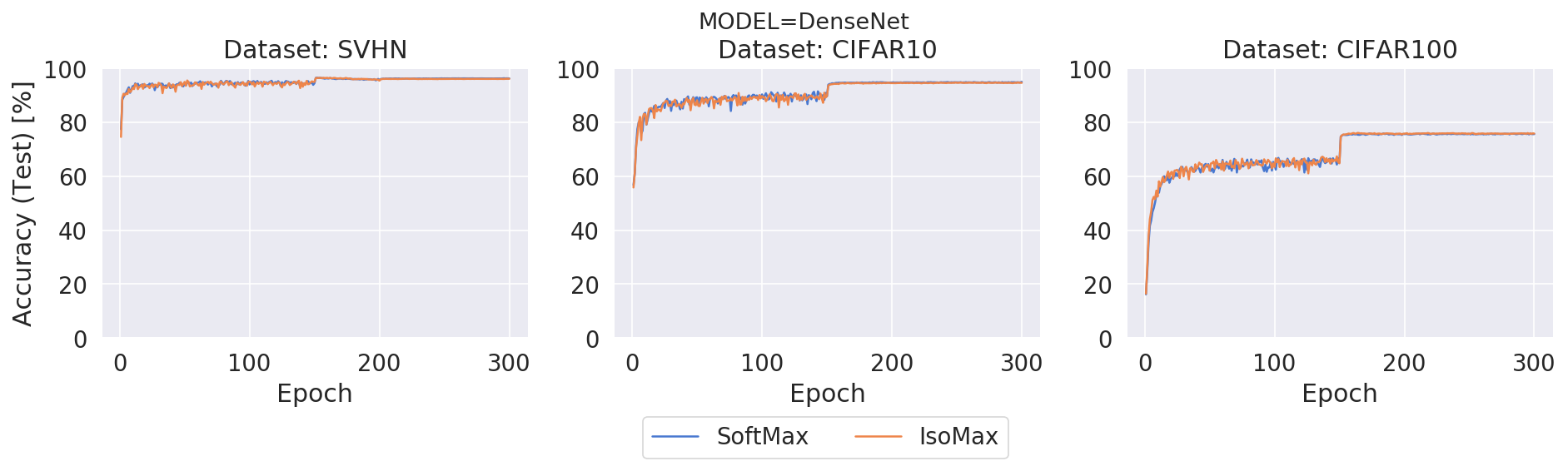}}
\\
\vskip -0.025cm
\subfloat[]{\includegraphics[width=\textwidth]{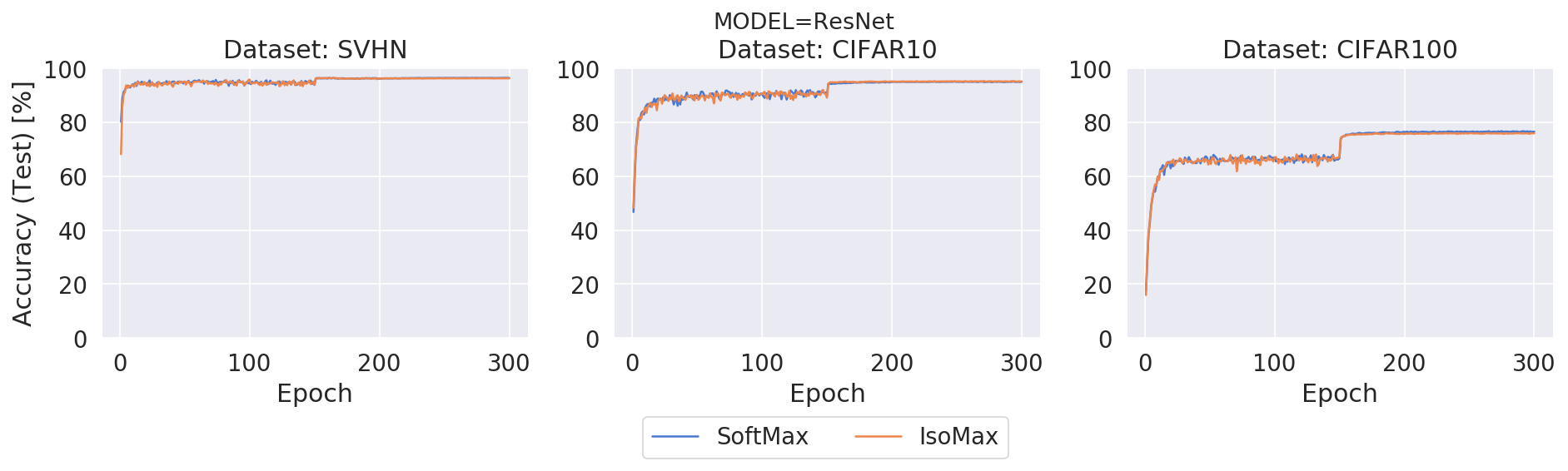}}
\\
\vskip -0.15cm
\caption{Training metrics: (a,b) Training loss values and (c,d) test accuracies for a range of models, datasets, and losses. SoftMax loss and IsoMax loss present remarkably similar metrics throughout training, which confirms that IsoMax loss training is consistent and stable.}
\label{fig:training_metrics}
\end{figure*}

\section{SoftMax Loss Anisotropy}\label{apx:softmax-loss-anisotropy}

Let $\bm{x}$ represent the input applied to a neural network and $\bm{f}_{\bm{\theta}}(\bm{x})$ represent the high-level feature vector produced by it. For this work, the underlying structure of the network does not matter.
Considering $k$ to be the correct class for a particular training example $\bm{x}$,
we can write the SoftMax loss associated with this specific training sample as:

%\begin{equation}
%\begin{multline}
\begin{align}
\label{eq:loss_softmax}
\mathcal{L}_{S}(\hat{y}^{(k)}|\bm{x})&=-\log\left(\frac{\exp(\bm{w}_k^\top\bm{f}_{\bm{\theta}}(\bm{x})\!+\!b_k)}{\sum\limits_j\exp(\bm{w}_j^\top\bm{f}_{\bm{\theta}}(\bm{x})\!+\!b_j)}\right)%\\+\mathcal{R}_{S}(\bm{f}_{\bm{\theta}}(\bm{x}),\bm{w},b)
%\end{multline}
\end{align}
%\end{equation}

In Equation~\eqref{eq:loss_softmax}, $\bm{w}_j$ and $b_j$ represent the weights and biases associated with class $j$, respectively. The sum has $N$ terms, where $N$ is the number of classes. From a geometric perspective, the term $\bm{w}_j^\top\bm{f}_{\bm{\theta}}(\bm{x})\!+\!b_j$ represents a hyperplane in the high-level feature space. It divides the feature space into two subspaces that we call positive and negative subspaces. The deeper inside the positive subspace the feature vector $\bm{f}_{\bm{\theta}}(\bm{x})$ of a particular example is located, the more likely the example is believed to belong to the considered class.

Therefore, training neural networks using SoftMax loss does not incentivize the agglomeration of representations of examples associated with a particular class into a limited region of the hyperspace, as it produces \emph{separable features} rather than \emph{discriminative features}~\cite[Fig.~1]{DBLP:conf/eccv/WenZL016}. The immediate consequence is the propensity of neural networks trained with SoftMax loss to make high confidence predictions on examples that stay in regions far away from the training examples, which explains their unsatisfactory OOD detection~performance~\cite{Hein2018WhyRN}. Indeed, the SoftMax loss is based on affine transformations, which are essentially internal products. Consequently, the last layer representations of such networks tend to align in the direction of the weight vector, producing locally preferential directions in space and subsequently~anisotropy.

The SoftMax loss anisotropy is usually corrected by using metric learning on neural network pretrained features \cite{lee2018simple, Mensink2013DistanceBasedIC, Scheirer_2013_TPAMI, Scheirer_2014_TPAMIb, 7298799,Rudd_2018_TPAMI}. For example, the high OOD detection performance of the Mahalanobis approach \cite{lee2018simple} indicates that deploying locally isotropic spaces around class prototypes improves the OOD detection performance. In such solutions, a mapping from the extracted features to a novel embedding space is constructed, and class prototypes are produced. The distance may be predefined (e.g., Euclidean distance) or learned (e.g., Mahalanobis distance).

However, approaches based on feature extraction and metric learning present drawbacks \cite{musgrave2020metric}. First, they are not turnkey, as additional procedures are required after neural network training. Additionally, they usually present hyperparameters to tune, typically requiring unrealistic access to design-time OOD or adversarial samples. \emph{Therefore, a possible option to build a seamless approach to OOD detection is to design an isotropic (exclusively distance-based) loss that works as a SoftMax loss drop-in replacement.}

\section{Nonsquared Euclidean Distance}\label{apx:euclidean-distance}

We need to choose a distance that allows IsoMax to work as a SoftMax drop-in replacement. Hence, the loss needs to learn \emph{both} high-level features \emph{and prototypes} using \emph{exclusively} SGD and \emph{end-to-end} backpropagation, as \emph{no additional offline procedures are allowed}. We also require the training using IsoMax loss to be \emph{as consistent and stable as the typical SoftMax loss neural network~training}. 

The covariance matrix makes it hard to use the Mahalanobis distance to train a neural network directly. Therefore, we decide to use \emph{Euclidean} distance. We have reasons to prefer the \emph{nonsquared} Euclidean distance rather than the \emph{squared} Euclidean distance. First, the \emph{nonsquared} Euclidean distance is a real metric that obeys the Cauchy–Schwarz inequality while the \emph{nonsquared} Euclidean distance is not. Using a metric that follows the Cauchy–Schwarz inequality is essential because of our previous geometric considerations. Additionally, using the \emph{squared} Euclidean distance is actually equivalent to using a linear model with a particular parameterization\cite{Snell2017PrototypicalNF, Mensink2013DistanceBasedIC}, which we prefer to avoid to increase the representative power of the proposed loss.

Moreover, we experimentally observed that training neural networks using \emph{exclusively} SGD and \emph{end-to-end} backpropagation is more stable and consistent when using \emph{nonsquared} Euclidean distance rather than \emph{squared} Euclidean distance. Indeed, \emph{squared} Euclidean distance-based logits are harder to \emph{seamlessly} train than \emph{nonsquared} Euclidean distance-based logits because numeric calculus instabilities are much more likely to occur when performing calculations and derivations with values of the order of $\mathcal{O}(e^{-d^{2}})$ than $\mathcal{O}(e^{-d})$. Finally, even in the cases in which we were eventually able to \emph{seamlessly} train neural networks using the \emph{squared} Euclidean distance, we observed lower OOD detection performance than using the \emph{nonsquared} Euclidean distance-based alternative. Therefore, we choose the \emph{nonsquared} Euclidean distance.

\section{Entropic Score Discussion}\label{apx:entropic_score}

Out-of-distribution detection approaches typically define a score to be used after inference to evaluate whether an example should be considered OOD. In a seminal work, \cite{Shannon1948ACommunication} demonstrated that the entropy presents the optimum measure of the randomness of a source of symbols. More broadly, we understand entropy as a measure of the uncertainty we have about a random variable. Considering that the uncertainty in classifying a specific sample should be an optimum metric to evaluate whether a particular example is OOD.

During IsoMax training, the embedding-prototype distances are affected. On the one hand, the distances from embeddings to the correct class prototype are reduced to increase classification accuracy. On the other hand, the distances from embeddings to the wrong class prototypes are increased. Consequently, based on the Equation~\eqref{eq:probability_isomax}, the probabilities of in-distribution examples increase. Therefore, it is reasonable to expect that samples with lower entropy more likely belong to the in-distribution.

From a practical perspective, using this predefined mathematical expression as score avoids the need to train an additional regression model to detect OOD samples that is otherwise required, for example, in the Mahalanobis approach.

Even more important, since no regression model needs to be trained, there is no need for unrealistic access to OOD or adversarial samples for hyperparameter validation. Since the entropic score is a predefined mathematical expression rather than a trainable model, it is available as soon as the neural network training finishes, avoiding additional training of extra models in a post-processing phase.

\section{Differentiation from the Scaled Cosine Approach}\label{apx:scaled_cosine}

An approach based on cosine similarity is proposed in \cite{techapanurak2019hyperparameterfree}. However, different from IsoMax loss, this solution presents \emph{significant classification accuracy drop} (the authors even suggest using two models, one for classification and the other for OOD detection) and increases the total number of parameters by requiring an additional layer.

Additionally, it is not an isotropic loss, as the cosine similarity is used rather than a distance. Moreover, the authors explicitly say that they do not have an explanation of why their solution works. Finally, it does not work as a SoftMax loss drop-in replacement, as we need to change the optimizer to avoid applying weight decay to the last layer.

\section{Robustness Study}\label{apx:robustness_study}

Fig.~\ref{fig:classification_robstness} shows that the classification accuracy of models trained with the IsoMax loss is similar to the models trained with the SoftMax loss regardless of the dataset or architecture. This is also true for a varying number of training examples per class.

Fig.~\ref{fig:detection_robstness} presents the OOD detection performance of SoftMax and IsoMax losses in many models (DenseNet and ResNet), and datasets (SVHN, CIFAR10, and CIFAR100) for a range of training samples per class. For each in-distribution, the AUROC and TNR@TPR95 results were averaged over all possible combinations of OOD detection test sets. The entropic score was used in all cases. It shows that IsoMax loss consistently and notably overcomes the OOD detection performance of SoftMax loss for virtually all combinations of datasets, training examples per class, metrics, and~models.

\section{Additional Analyses}\label{apx:additional_analyses} % MOVED FROM MAIN PAPER!!!

Fig.~\ref{fig:logits_histograms} shows that the in-distribution \emph{interclass} logits are more distinguishable from out-distribution logits when using IsoMax loss, which explains its increased OOD detection performance compared with the SoftMax loss since there are far more \emph{interclass} logits than \emph{intraclass} logits. Distances are calculated from class prototypes.

The unimodal nature of the out-distribution can also help to explain how IsoMax can distinguish between in-distributions and out-distributions. Indeed, for out-distributions examples, we do not distinguish between intraclass and interclass logits, as there is no such thing that we can call the correct class in such cases. Therefore, on the one hand, we can see that the out-distributions logits are all very closed clustered (green distributions), producing even higher entropies than in-distributions examples. On the other hand, intraclass logits (blue distributions) are apart from interclass logits (orange distributions), generating high but slightly smaller entropies.

SoftMax intraclass and interclass logits are much farther apart than the IsoMax intraclass and interclass logits. Consequently, in disagreement with the maximum entropy principle, the former produces much lower entropies than the latter for in-distribution data.

Fig.~\ref{fig:maxprobs_entropies_histograms} shows that networks trained with SoftMax loss exhibit high maximum probabilities. Sometimes this is true even for OOD samples. For networks trained with IsoMax loss, OOD samples usually present lower maximum probabilities compared with in-distribution samples. Furthermore, it also shows that the networks trained with SoftMax loss are overconfident.

We reemphasize that this work aims to perform OOD detection. Unlike uncertainty/calibration estimation/calibration approaches, we do not claim or intend that the probabilities produced are calibrated. To agree with the principle of maximum entropy, rather than ``plausible probabilities'', we need to produce the lowest possible probabilities that are, nevertheless, capable of providing the highest classification accuracy possible. We may even question the meaning of calibrated probabilities in an open set environment such as OOD tasks, where we do not even know how often the system will deal with OOD examples.

Additionally, Fig.~\ref{fig:maxprobs_entropies_histograms} shows that the entropy works as a high-quality score to distinguish the in-distribution from the out-distribution in neural networks trained with IsoMax loss.

\section{Training Metrics}\label{apx:training_metrics}

Fig.~\ref{fig:training_metrics} shows that the training metrics are remarkably similar for SoftMax and IsoMax losses. Regardless of the dataset and model used, the loss values produced are similar. The same is true for the validation accuracy along the network training.

% that is all folks

\end{document}